\documentclass[journal]{IEEEtran}

\usepackage{tabularx,booktabs,multirow}
\usepackage[table]{xcolor}
\usepackage{colortbl}
\definecolor{myorange}{HTML}{8497B0}
\definecolor{myyellow}{HTML}{FFC000}
\definecolor{mygreen}{HTML}{00B050}
\definecolor{myblue}{HTML}{4472C4}

\usepackage{ifpdf}
  \ifpdf
  \else
  \fi

\usepackage{cite}

\ifCLASSINFOpdf
  \usepackage[pdftex]{graphicx}
\else
\fi

\usepackage{amsmath}
\usepackage{soul, color, xcolor}

\soulregister{\cite}7
\soulregister{\ref}7

\usepackage{algorithmic}
\usepackage{algorithm}

\usepackage{array}

\ifCLASSOPTIONcompsoc
  \usepackage[caption=false,font=normalsize,labelfont=sf,textfont=sf]{subfig}
\else
  \usepackage[caption=false,font=footnotesize]{subfig}
\fi

\usepackage{stfloats}
\ifCLASSOPTIONcaptionsoff
  \usepackage[nomarkers]{endfloat}
 \let\MYoriglatexcaption\caption
 \renewcommand{\caption}[2][\relax]{\MYoriglatexcaption[#2]{#2}}
\fi

\usepackage{url}

\begin{document}

% Title
\title{One-Class Knowledge Distillation for Face Presentation Attack Detection}

% Authors
\author{Zhi Li,
        Rizhao Cai,
        Haoliang Li,
        Kwok-Yan Lam,
        Yongjian Hu,
        and Alex C. Kot~\IEEEmembership{Fellow,~IEEE}% stops a space

\thanks{Zhi Li and Kwok-Yan Lam are with the School of Computer Science and Engineering, Nanyang Technological University, Singapore.

Rizhao Cai is with the School of Electrical and Electronic Engineering, Nanyang Technological University, Singapore.

Haoliang Li (Corresponding Author) is with the Department of Electrical Engineering, City University of Hong Kong, Hong Kong.

Yongjian Hu is with the School of Electronic and Information Engineering, South China University of Technology, Guangzhou, China, and with China-Singapore International Joint Research Institute.

Alex C. Kot is with the School of Electrical and Electronic Engineering, Nanyang Technological University, Singapore, and with China-Singapore International Joint Research Institute.

}
}
\maketitle
\begin{abstract}
Face presentation attack detection (PAD) has been extensively studied by research communities to enhance the security of face recognition systems. Although existing methods have achieved good performance on testing data with similar distribution as the training data, their performance degrades severely in application scenarios with data of unseen distributions. In situations where the training and testing data are drawn from different domains, a typical approach is to apply domain adaptation techniques to improve face PAD performance with the help of target domain data. However, it has always been a non-trivial challenge to collect sufficient data samples in the target domain, especially for attack samples. This paper introduces a teacher-student framework to improve the cross-domain performance of face PAD with one-class domain adaptation. In addition to the source domain data, the framework utilizes only a few genuine face samples of the target domain. Under this framework, a teacher network is trained with source domain samples to provide discriminative feature representations for face PAD. Student networks are trained to mimic the teacher network and learn similar representations for genuine face samples of the target domain. In the test phase, the similarity score between the representations of the teacher and student networks is used to distinguish attacks from genuine ones. To evaluate the proposed framework under one-class domain adaptation settings, we devised two new protocols and conducted extensive experiments. The experimental results show that our method outperforms baselines under one-class domain adaptation settings and even state-of-the-art methods with unsupervised domain adaptation.
\end{abstract}

\ifCLASSOPTIONpeerreview
\begin{center}
\bfseries EDICS Category: 3-BBND
\end{center}
\fi
\IEEEpeerreviewmaketitle

\section{Introduction}
\begin{figure}[htbp]
\centering
\includegraphics[width=0.4\textwidth]{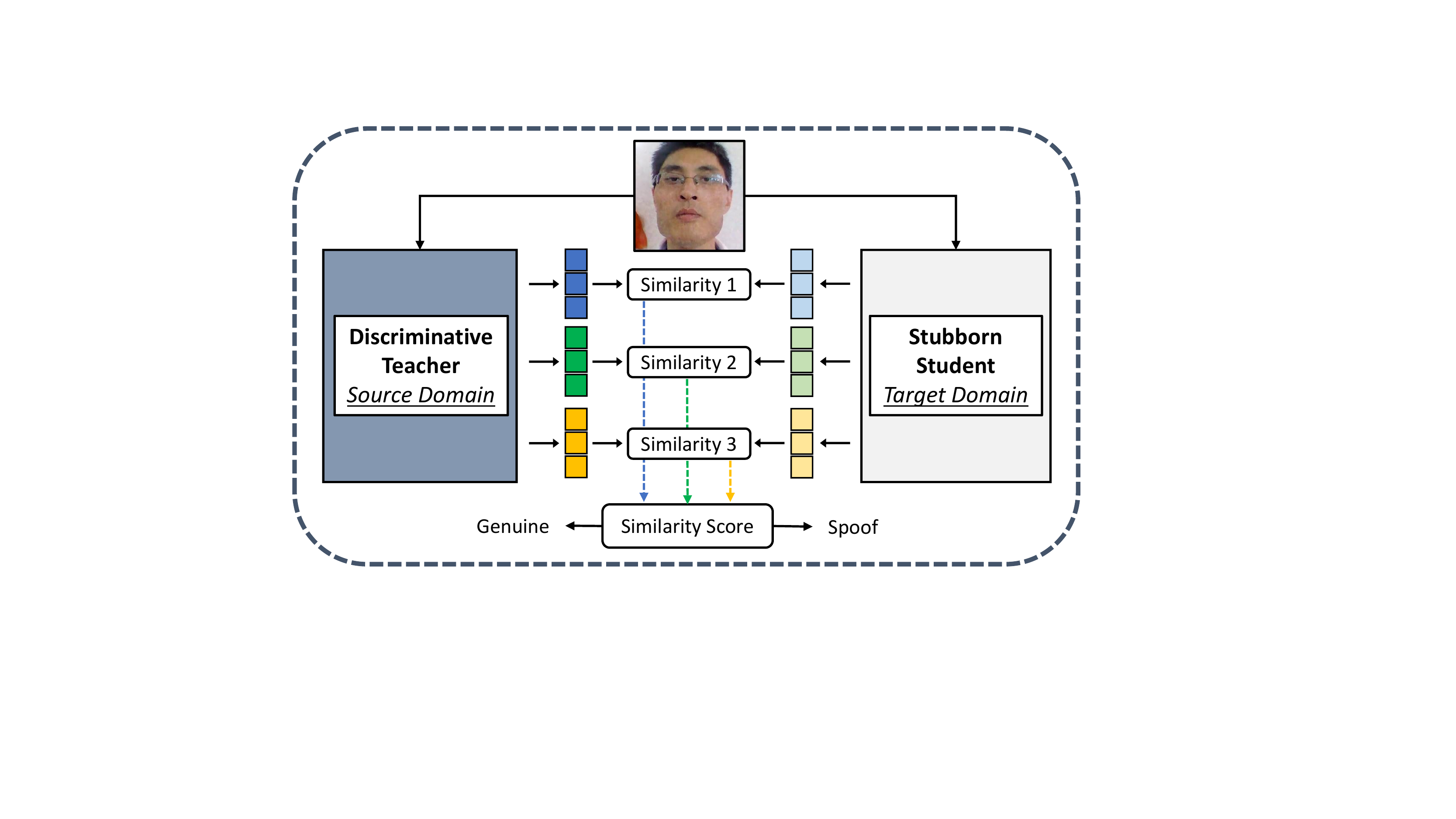}
\caption{The framework contains a teacher network trained with the source domain data to provide discriminative features, and a student network trained with the target domain genuine face data to generate similar features to the teacher's descriptions. In the test phase, the face images will be fed into both the DT and SS netorks for feature extraction and the similarity between features of the two networks will be used as the inference score.}
\vspace{-0.5cm}
\label{fig:figure_01}
\end{figure}
\IEEEPARstart{F}{ace} recognition as a convenient automatic identity verification approach has become increasingly prevailing in recent years. With the wide application of face recognition techniques in various scenarios, the security and trustworthy problems of face recognition systems have been widely concerned in academia and industry. On account of the easy deployment, presentation attack (PA), also known as spoofing attack, is a crucial threat hindering the application of face recognition systems. Face presentation attack aims to spoof the face recognition systems and be verified as the attempts of genuine users. As the name implies, it is conducted by presenting facial forgery (printed photos, images or videos on digital displays, high-fidelity masks, expensive wax figures, etc.) of genuine users to the camera sensor of face recognition systems at the sensing process. 

Face presentation attack detection (PAD) is an essential anti-spoofing measurement to enhance the security and reliability of face recognition systems by discriminating presentation attacks from bona fide attempts. In the past decade, various face PAD methods have emerged in the literature, ranging from the early traditional methods based on handcrafted features \cite{LBP_ICIP15, HARALICK_BTAS16, IQA_TIP14, IDA_TIFS15} to recent deep learning methods \cite{CNN_Arxiv14, BA_CVPR18, Revisiting_TBIO21, RF_Depth_BM_ECCV20, CDCN_CVPR20, LBP_ECCV20, IDR_TIST20, BM_TIFS20, Denoising_ECCV18, ODPT_ECCV20, SRA_TIFS20, SFS_TIP20, ACMT_TIFS21, AMT_TMM21, DRL_TIFS20, NASFAS_TPAMI21, LWN_ICASSP20, ANOMALY_ACCESS17, HYPER_ICASSP20, LLIG_TIFS20, CSAD_PR21, MK_TIFS21, LDGDFR_TIFS18, MDDRL_CVPR20, VSA_MM21, DRDG_IJCAI21, LMM_AAAI20, RFM_AAAI20, ANRL_MM21, MT_TPAMI21, GRLMD_AAAI21, KSA_TIFS18, ADA_ICB19, FASDNND_SP20, USDAN-Un_PR21, UDA_TIFS20, SDA_TIFS15, DGP_ICASSP20, OCA-FAS_NC20, MP_TIFS22}. Existing methods have achieved good performance in intra-domain testing, where the testing data is from the same distribution as training data. However, when testing the face PAD models in a new target domain, the performance will degrade severely since the testing data is from unseen distributions which are different from the training data. This problem is also known as the distribution shift or domain shift problem, which originates from various factors such as the change of capturing devices, mediums of attacks, and illumination \cite{OULU, competition}.

Since the domain shift seriously affects the reliability of face PAD models, domain adaptation techniques have been used to address the cross-domain problems in face PAD recently. Domain adaptation techniques improve the cross-domain performance by utilizing the target domain data. However, it is difficult and expensive to collect and annotate sufficient data samples in the target domain for the adaptation. Moreover, collecting attack samples requires facial forgeries. The production of facial forgeries is complicated, and there is no guarantee that the collected attack samples are the same as those launched by attackers. Compared to attack samples, genuine face samples are much easier and cheaper for collection. Therefore, we expect to improve the cross-domain performance of the face PAD model by only using a few genuine face samples collected in the target domain, which is named as One-Class Domain Adaptation (OCDA) \cite{OCA-FAS_NC20}. The straightforward approach for OCDA is to train a face PAD model with the source domain data and fine-tune the pre-trained model with the target domain data. However, such a naive fine-tuning approach cannot provide good performance due to the one-class characteristic of the target domain data. Recently, Qin \textit{et al.} \cite{OCA-FAS_NC20} propose a meta-learning framework, which incorporates a meta loss function search (MLS) strategy to search for better loss functions and help the meta-learner deal with the OCDA tasks. Mohammadi \textit{et al.} \cite{DGP_ICASSP20} propose a framework to tackle the OCDA tasks with domain guided pruning. Under the framework, the generalization ability of different filters in the pre-trained face PAD model is estimated with the feature divergence between genuine face samples from the source and the target domain, and the filters with poor generalization ability are pruned. Besides, Fatemifar \textit{et al.} \cite{CSAD_PR21} propose to develop client-specific face PAD models with some genuine face samples collected in the target client domain. Pre-trained deep neural networks are used as feature extractors to build face PAD models with traditional one-class classifiers. Recent works show that it is promising to facilitate the training of face PAD models with some genuine face samples collected in the target domain. However, the performance of existing methods needs to be further improved, especially in scenarios where mixture factors cause a large distribution shift between the source and the target data domain. How to leverage the genuine face samples to effectively improve the target domain performance of face PAD is still a crucial problem to be studied.

\begin{figure}[tbp]
\centering
\includegraphics[width=0.4\textwidth]{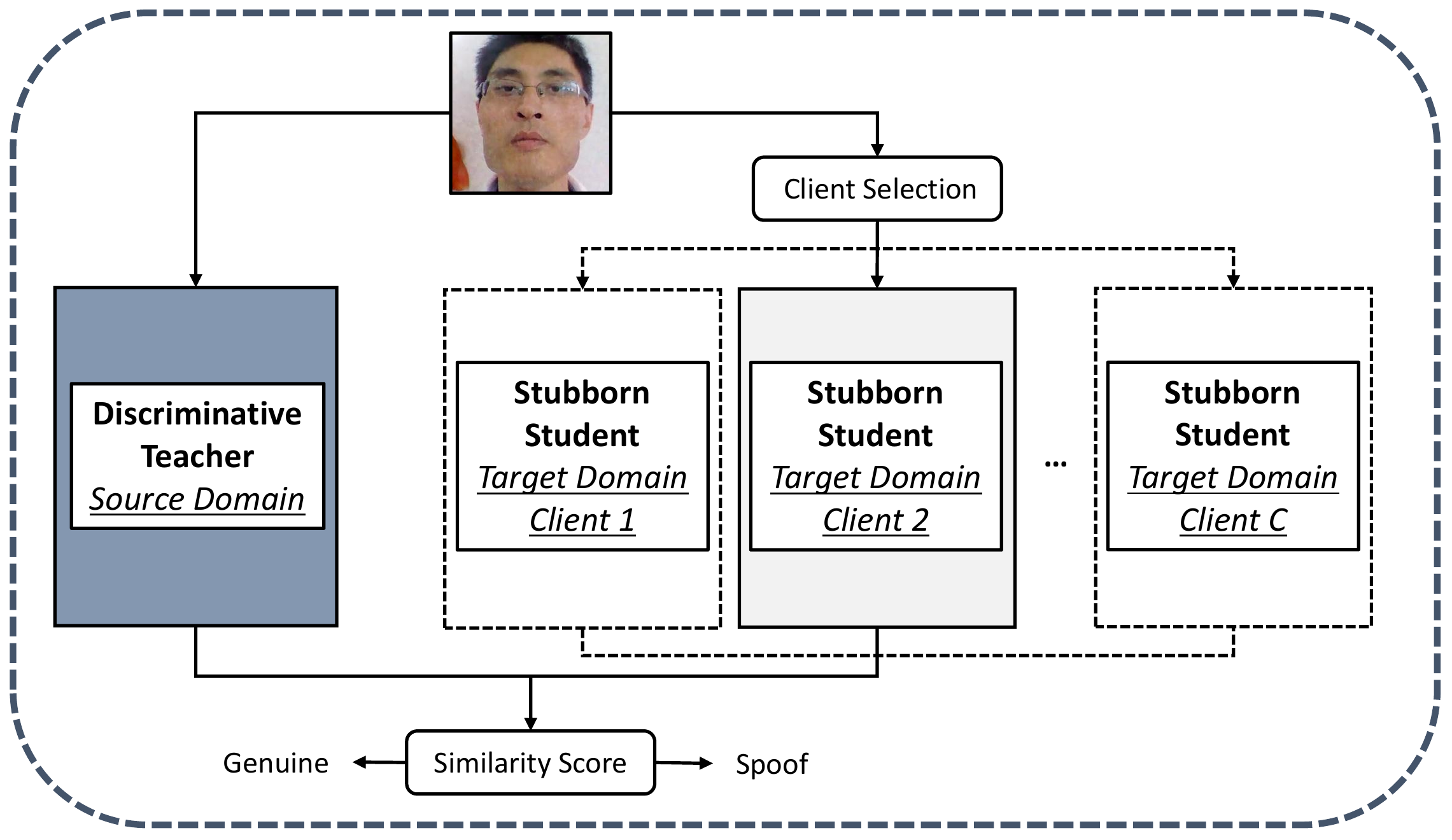}
\caption{After the client-specific one-class domain adaptation, the framework contains a teacher network and a set of $N$ student networks. Each student network serves for one specific target client. In the test phase, the face images will be fed into the DT network and the corresponding SS network for client-specific inference.}
\vspace{-0.8cm}
\label{fig:figure_02}
\end{figure}
This work tackles the OCDA problem in face PAD with a teacher-student framework. Different from previous teacher-student frameworks for model compression and fine-grained supervision \cite{FASDNND_SP20, MT_TPAMI21}, our framework is formulated in the anomaly detection paradigm to tackle the OCDA problem. The insight behind such formulation is that the target domain data for training contains only genuine face samples, and the attacks encountered during the deployment can be regarded as anomalies. In our framework, a teacher network is trained with the source domain data to provide multi-level discriminative feature representations for face PAD. Inspired by anomaly detection \cite{US_CVPR20}, a student network is trained with only genuine face samples of the target domain to generate similar representations to the teacher's outputs. As such, the student network is ``stubborn” and only learns similar representations for the genuine face images. Therefore, the genuine face representations of the teacher and student networks are more similar than the spoof ones. Finally, we use the similarity score to discriminate attacks from the genuine ones, as illustrated in Fig. \ref{fig:figure_01}. Moreover, recent literature \cite{SDA_TIFS15, CSAD_PR21} points out that different target clients could be reckoned as different client-specific domains. Therefore, we also consider the OCDA problem under a client-specific setting. Client-Specific One-Class Domain Adaptation (CS-OCDA) aims to develop a specific model for each target client with a few of its own genuine face samples. Our proposed framework is flexible to be extended for the CS-OCDA by training a ``stubborn'' student model for each client. In CS-OCDA, although the use of the student networks multiplies the number of parameters as the number of clients increase, we apply a sparse training strategy during the optimization to shrink the size of student networks and alleviate the storage problem.

Our main contributions in this work are summarized as below:
\begin{itemize}
\item We introduce a one-class knowledge distillation framework to address the cross-domain problem in face PAD, which improves the target domain performance by utilizing only a few genuine face samples in the target domain.

\item We devise two new protocols on public benchmark datasets for the performance evaluation of face PAD methods under the general and client-specific one-class domain adaptation settings.

\item We conduct extensive experiments to verify the effectiveness of our proposed method. The experimental results show that our proposed method outperforms baseline methods under one-class domain adaptation settings and even state-of-the-art methods with unsupervised domain adaptation.
\end{itemize}
\section{Related Works}
Over the past decade, face PAD methods have evolved rapidly from traditional methods based on handcrafted features to deep learning methods. Recently, the cross-domain problems of face PAD have become a research hot-spot. Domain generalization and adaptation methods have been studied to improve the performance of face PAD models in the target domain. In the following content of this section, we will firstly introduce the evolution of face PAD methods. After which, the review will be focused on the face PAD methods based on domain generalization and domain adaptation, which are most relevant to our work.

\subsection{Evolution of Face PAD Methods}
Face presentation attack detection is a task to discriminate spoofing attacks from the attempts of genuine users. Although diverse camera sensors have been used for information acquisition, RGB image based methods form the main branch of face PAD research due to their convenience and low cost. At the earlier stage, a variety of handcrafted features \cite{LBP_ICIP15, HARALICK_BTAS16, IQA_TIP14, IDA_TIFS15} have been devised to represent the data samples in the discriminative feature space, which is easier for classifiers to distinguish between genuine and spoof face images. These handcrafted features are designed by domain experts on the basis of specific domain knowledge such as textures analysis \cite{LBP_ICIP15, HARALICK_BTAS16}, and image quality analysis \cite{IQA_TIP14, IDA_TIFS15}. The representational ability of the handcrafted features is usually limited by prior knowledge about specific attacks, and rare of these features are widely effective for the discrimination between genuine samples and various types of attacks.

With the great success of deep learning techniques in the field of representation learning and various computer vision tasks, deep neural networks have been used for face PAD and demonstrated significant advantages and great potentials. VGG-Net \cite{VGG_Arxiv14} has been used as the feature extractor to facilitate the classification of genuine and spoofing face images in \cite{CNN_Arxiv14}. This is the first attempt at using deep learning techniques on the face PAD task. After which, an increasing number of deep learning based face PAD methods have emerged along with the rapid development of deep learning techniques. The good performance of deep learning based methods relies on the design of neural network architectures and the optimization objectives, as well as the diversity of training data and the quality of annotations. Since the plain neural networks naively optimized with binary class labels can hardly satisfy the performance requirements, various types of auxiliary tasks such as depth \cite{BA_CVPR18, Revisiting_TBIO21, RF_Depth_BM_ECCV20, CDCN_CVPR20}, reflection \cite{RF_Depth_BM_ECCV20}, texture map \cite{LBP_ECCV20}, and binary masks\cite{BM_TIFS20} generation, noises \cite{Denoising_ECCV18} and spoofing traces \cite{ODPT_ECCV20} modeling, intrinsic image decomposition \cite{SRA_TIFS20, SFS_TIP20}, and modality translation \cite{ACMT_TIFS21, AMT_TMM21} have been designed to assist the training of deep neural networks for face PAD. Besides, deep reinforcement learning has been used to model the behavior of exploring spoofing cues from sub-patches of images in \cite{DRL_TIFS20}. Network architecture search (NAS) techniques have been used to automatically search the optimal parameters of network architectures for higher accuracy \cite{CDCN_CVPR20, NASFAS_TPAMI21} and efficiency \cite{LWN_ICASSP20}. Since the attack data samples are more difficult for collection and the types of attack in practical applications are uncontrollable, a series of anomaly detection based face PAD methods \cite{ANOMALY_ACCESS17, HYPER_ICASSP20, LLIG_TIFS20, CSAD_PR21, MK_TIFS21} have been proposed to address the face PAD problem under unseen attack settings.

Despite existing data-driven face PAD methods performing well on intra-domain testing data, the excellent performance is limited by the diversity of the source domain training data, and seldom of them can generalize well on the target domain due to the domain shift problem \cite{KSA_TIFS18} caused by complex factors.

\subsection{Domain Generalization and Domain Adaptation for Face PAD}
Since the domain shift problem severely interferes with the reliability of face PAD models, domain generalization and domain adaptation techniques have been applied in the face PAD task to improve the target domain performance in recent years.

Domain generalization based methods assume that the target domain data is unavailable for the model training and aim to develop a generalized model by utilizing data samples from multiple source domains. Li \textit{et al.} \cite{LDGDFR_TIFS18} design a generalization loss that guides the neural network to learn generalized feature representation by manipulating the feature distribution distances of different data domains. Disentangled representation learning techniques \cite{MDDRL_CVPR20, VSA_MM21} have been used to learn domain-independent features for generalized face PAD models. Wang \textit{et al.} \cite{MDDRL_CVPR20} propose a framework with a disentangled representation learning module (DR-Net) and a multi-domain learning module (MD-Net) to force the feature representations of the face PAD model to more subject-independent and domain-independent. Wang \textit{et al.}\cite{VSA_MM21} argue that using simple global pooling makes the representations of face PAD models lose local feature discriminability. Therefore, they propose a framework based on a modified vector of locally aggregated descriptors (VLAD) to learn better feature representations. Considering the complex and biased distribution of different data domains, Liu \textit{et al.} \cite{DRDG_IJCAI21} propose a framework that iteratively reweights the relative importance between samples to further improve the generalization. Besides, there are several meta-learning frameworks \cite{LMM_AAAI20, RFM_AAAI20, ANRL_MM21, MT_TPAMI21} have been proposed to learn generalized face PAD models with specific meta tasks under the simulated train and test scenarios with domain shifts. To tackle the problem that face PAD models are easy to overfit on seen attacks, Qin \textit{et al.} \cite{LMM_AAAI20} propose a meta-learning framework with a Fusion Training (FT) and Adaptive Inner-Update (AIU) learning rate strategy. In \cite{RFM_AAAI20}, Shao \textit{et al.} propose a different meta-learning framework to address the domain generalization problem. The framework simultaneously conducts meta-learning in multiple scenarios with simulated domain shifts and incorporates a regularization based on domain knowledge. Liu \textit{et al.} \cite{ANRL_MM21} argue that the normalization of features has a great impact on the generalization of the learned representation. To learn more generalizable representations for face PAD, they propose a meta-learning framework with an Adaptive Feature Normalization Module for normalization method selection and a Dual Calibration Constraints to direct the model optimization. To explore better supervision for face PAD, Qin \textit{et al.} \cite{MT_TPAMI21} propose a bi-level optimization framework with a meta-teacher to provide better-suited supervision for the PA detectors to learn rich spoofing cues. Most domain generalization based methods require using the domain labels during training, while manually assigning training data with different domain labels is expensive, and the partition is usually sub-optimal. To avoid this limitation, Chen \textit{et al.} \cite{GRLMD_AAAI21} propose a method without using domain labels, which iteratively divides mixture domains under a meta-learning framework.

As complementary to the domain generalization methods, domain adaptation techniques have also been used for face PAD to further improve the target domain performance with some target domain training data. Most of the existing works formalize the face PAD problem under the unsupervised domain adaptation setting and aim to improve the performance of face PAD models with unlabelled target domain data. Li \textit{et al.} \cite{KSA_TIFS18} propose a framework to learn a more generalized classifier for face PAD by minimizing the Maximum Mean Discrepancy between the latent features in the source and the target domains. Wang \textit{et al.} \cite{ADA_ICB19} propose an approach to improve the generalization capability of PAD via adversarial domain adaptation. Li \textit{et al.} \cite{FASDNND_SP20} propose a method based on the concept of neural network distilling. Jia \textit{et al.} \cite{USDAN-Un_PR21} propose a method to minimize the distribution discrepancy between the source and the target domains by performing marginal and conditional distribution alignment. Wang \textit{et al.} \cite{UDA_TIFS20} propose a disentangled representation approach to improve the generalization capability of PAD into new scenarios. Due to the difficulty and expense of data collection, collecting sufficient data samples in the target domain for the model training is unrealistic. Compared to attack samples, genuine face samples are much easier and cheaper for collection. 

Recently, some works aim to improve the performance of the face PAD model with only a few genuine face images of the target domain. The first attempt of using genuine face image samples to improve the target domain performance in face PAD dates back in \cite{SDA_TIFS15}, where Yang \textit{et al.} propose a method about virtual feature generation to address the problem that generic face PAD classifier based on handcrafted features cannot generalize well to all subjects. Recently, the concept of one-class domain adaptation has been used to improve the performance of face PAD models under the scenarios with domain shifts between the source and the target domains. Mohammadi \textit{et al.} \cite{DGP_ICASSP20} propose a method that uses domain guided pruning to adapt a pre-trained PAD network to the target dataset. Qin \textit{et al.} \cite{OCA-FAS_NC20} propose a meta-learning framework with a meta loss function search strategy to improve the performance of the face PAD model under the one-class domain adaptation setting. 

Similar to \cite{DGP_ICASSP20, OCA-FAS_NC20}, our work aims to improve the performance of face PAD by utilizing only genuine face samples of the target domain. But different from them, we propose a method based on one-class knowledge distillation, which applies to general and client-specific one-class domain adaptation settings. The detailed illustration of the methodology and the numerical comparison with relevant methods can be found in Section \ref{section: method} and Section \ref{section: experimental_results}, respectively.

\section{Methodology\label{section: method}}

\subsection{Problem Formulation and the Proposed Framework}
Existing face PAD models trained with the source domain data can not generalize well to the target domain data. Although domain adaptation techniques could improve the cross-domain performance of face PAD with the help of the data collected in the target domain, the data collection is expensive and complicated, especially for attack ones. Therefore, we expect to address the cross-domain problem in face PAD with one-class domain adaptation (OCDA). In addition to source domain training data, we aim to improve the cross-domain performance of face PAD by only using a few genuine face samples collected in the target domain. We formulate the OCDA problem of face PAD under both the general and client-specific settings. For general one-class domain adaptation, different target clients are reckoned as one general domain, and the objective is to develop a general model for all clients. For client-specific one-class domain adaptation, different target clients are reckoned as different client-specific domains. The objective is to develop a set of client-specific models. Each model serves one specific target client.

Considering the one-class characteristic of the target domain data, we treat the one-class domain adaptation of face PAD as an anomaly detection problem. Inspired by the advances in anomaly detection tasks \cite{US_CVPR20}, we introduce a teacher-student framework named one-class knowledge distillation framework. The framework applies to both the general and client-specific OCDA settings. Under the framework, a teacher network $\theta_{DT}$ is trained with the genuine and attack samples from the source domain $D_{src}$ to provide multi-level discriminative feature representations for face PAD. For the general one-class domain adaptation setting, a student network $\theta_{SS}$ is trained with only genuine data from the general target domain $D_{tgt}$ to generate similar representations to the teacher’s outputs. As such, the genuine face representations of the teacher and student networks are more similar than the spoof ones. In the test phase, we use the similarity score to detect attacks, as illustrated in Fig. \ref{fig:figure_01}. For client-specific one-class domain adaptation setting, a set of stubborn student networks $\theta_{SS}^{c}$ are trained with genuine data from multiple client-specific target domains $D_{tgt}^{c}$, each student network serves for a specific target client, as illustrated in Fig. \ref{fig:figure_02}. Although the use of the student networks multiplies the number of parameters, a sparse learning strategy is adopted during the training of the student networks to alleviate the expansion of the model size. 
In the following contents of this section, we'll introduce the design details and training strategies about the teacher and student networks.
\begin{algorithm}[t]
\caption{Training of the Discriminative Teacher Stream}
\label{algo_01}
\begin{algorithmic}[1]
\REQUIRE Source domain training data $D_{src}$, learning rate $\alpha_{1}$, maximum training iteration $K_{1}$, and batch size $N_{1}$.
\ENSURE DT parameters $\theta_{DT}$.
\STATE Initialize the DT parameters as $\theta_{DT}$ and the FCB parameters as $\theta_{FCB}$.
\FOR {$k=1$ \TO $K_{1}$}
\STATE Sample $N_{1}$ samples $x_{i}$ with the labels $y_{i}$ from $D_{src}$.
\STATE Extract multi-level features $f_{i}^{1}$, $f_{i}^{2}$, and $f_{i}^{3}$ by encoding $x_{i}$ with $\theta_{DT}$.
\STATE Predict the pixel map $d_{i}$ by processing $f_{i}^{1}$, $f_{i}^{2}$, and $f_{i}^{3}$ with $\theta_{FCB}$.
\STATE Compute $\mathcal{L}_{DT}$ with $d_{i}$ and $y_{i}$ as in Eq. (1).
\STATE Update $\theta_{DT}$ and $\theta_{FCB}$ with $\mathcal{L}_{DT}$, $\alpha_{1}$.
\ENDFOR

\RETURN $\theta_{DT}$
\end{algorithmic}
\end{algorithm}
\subsection{Discriminative Teacher Stream\label{section: method_DTS}}
\begin{figure*}[ht]
\centering
\includegraphics[width=0.9\textwidth]{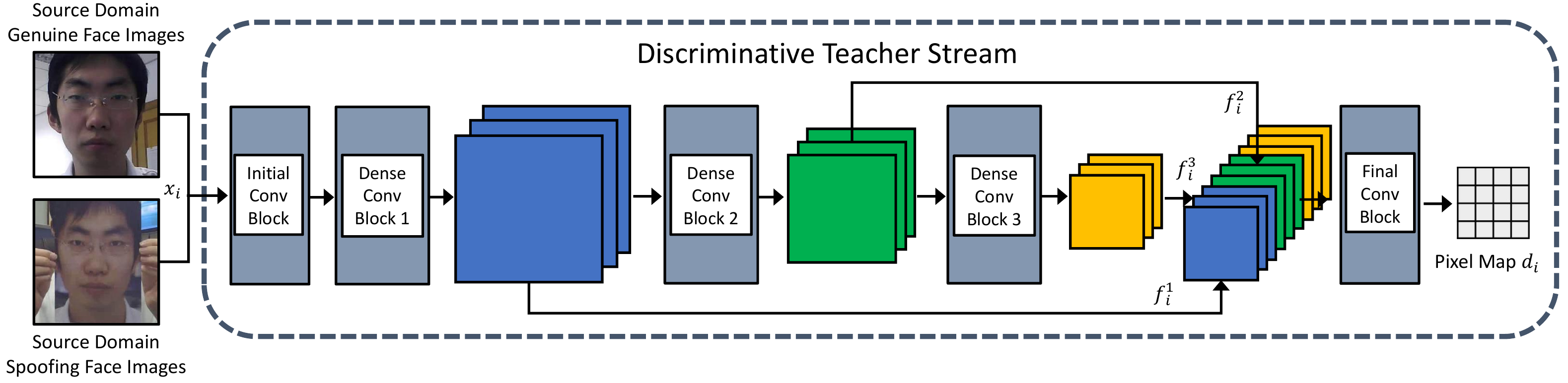}
\caption{The figure illustrates the network training of the Discriminative Teacher network with the source domain data. The face image $x_{i}$ will be fed into several convolutional blocks to extract features $f^{1}_{i}$, $f^{2}_{i}$ and $f^{3}_{i}$. The multi-level features are fed into the final convolutional block for pixel map $d_{i}$ estimation.}
\label{fig:figure_03}
\end{figure*}

The function of the Discriminative Teacher (DT) is to provide discriminative feature representations for the face PAD task. Benefit by the strong representational ability of deep learning, convolutional neural networks are widely used as the feature extractors for face PAD. We use the feature extractor of the depth regression network proposed in \cite{BA_CVPR18} as the backbone of our DT. A series of convolutional blocks are used to encode image $x_{i}$ into 3 level feature representations $f_{i}^{1}$, $f_{i}^{2}$, and $f_{i}^{3}$. In addition, a final convolutional block takes the multi-level features to estimate the pixel map $d_{i}$. To avoid the inconvenience of the pseudo depth map generation, we use binary maps $y_i$ with 0/1 value as the target of the pixel map as in \cite{DeepPixBis_ICB19}. Binary cross-entropy is used for loss calculation with $d_i$ and the target $y_i$ at the pixel-level, which is represented as
\begin{equation}
\begin{aligned}
\mathcal{L}_{DT} & = \frac{1}{N_{1}} \sum_{i=1}^{N_{1}} -(y_{i}\log(d_{i})+(1-y_{i})\log(1-d_{i})),\\
y_{i} & =\left\{
\begin{aligned}
0, &     & genuine,\\
1, &     & attack.
\end{aligned}
\right.
\end{aligned}
\label{eq:equation_01}
\end{equation}
If we use $\theta_{DT}$ and $\theta_{FCB}$ to denote the parameters of the feature extractor and the final convolutional block, the optimization problem is represented as,
\begin{equation}
\label{eq:equation_02}
\mathop{\arg\min}_{\theta_{DT}, \theta_{FCB}} \ \ \ 
E_{x,y \sim D_{src}} 
\mathcal{L}_{DT}(x, y|\theta_{DT}, \theta_{FCB}).
\end{equation}
Fig.\ref{fig:figure_03} illustrates the framework of the DT stream. The Algorithm \ref{algo_01} describes the details about the training of the DT. The DT is trained with genuine and attack samples from $D_{src}$. Since the expected function of the teacher stream is to provide multi-level discriminative features, the final convolutional block is abandoned after the DT network training.
\begin{figure*}[htbp]
\centering
\includegraphics[width=0.9\textwidth]{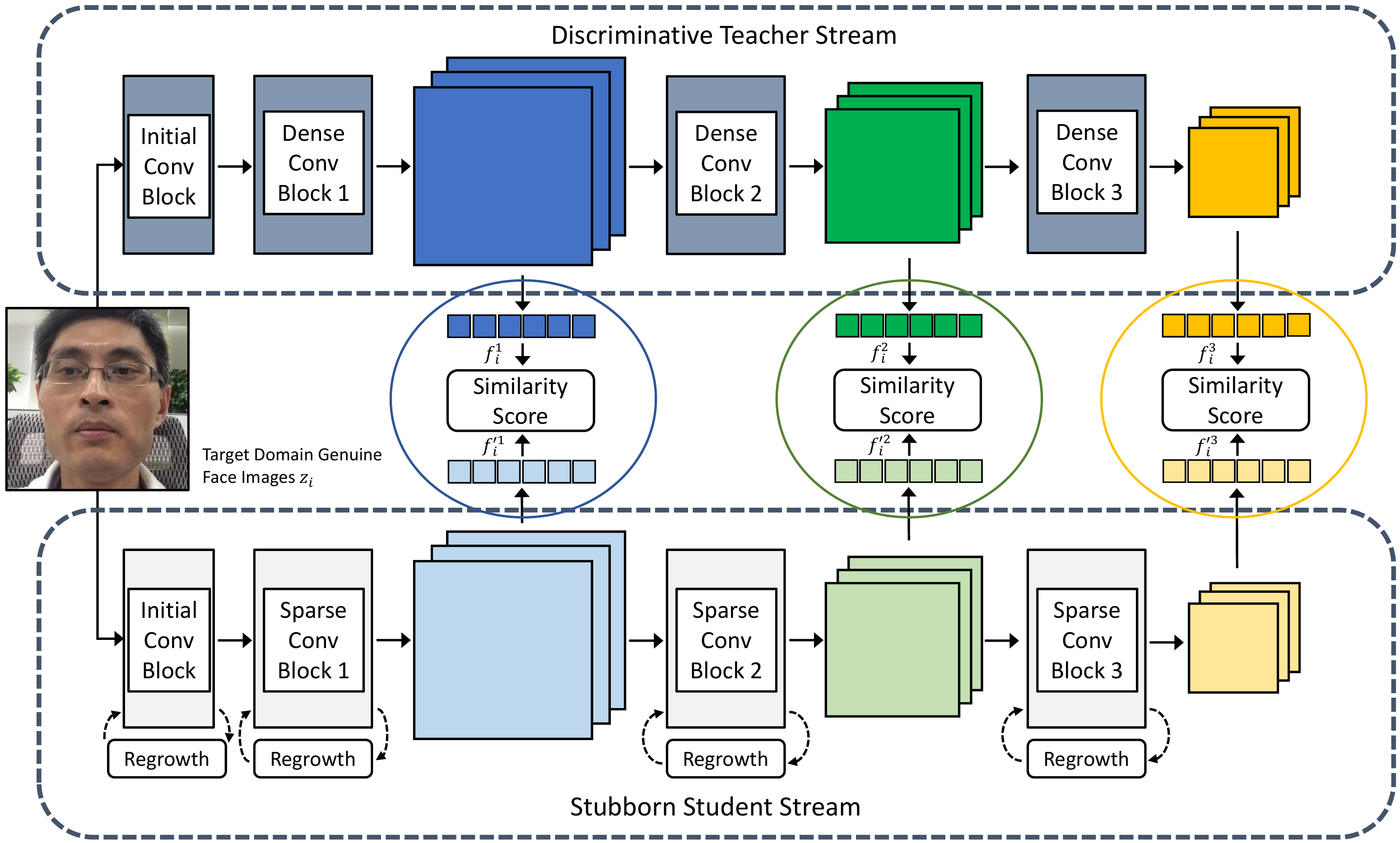}
\caption{The figure illustrates the training of the Stubborn Student network. The target domain genuine face images $z_{i}$ will be fed into both the teacher and student networks for feature extraction. The multi-level similarities between the DT and SS networks are computed to guide the optimization of the SS network. During the training, a sparse training strategy is used to reduce the parameter density of the SS network.}
\label{fig:figure_04}
\end{figure*}
\subsection{Stubborn Student Stream}
The learning objective of the Stubborn Student (SS) is to generate similar genuine face representations to the teacher's descriptions. We expect the student network is inflexible so that the representations of the student network are similar to the teacher's representations only for the genuine face images. During the training, genuine face images $z_{i}$ from the target domain $D_{tgt}$ are fed into both the DT and SS for feature extraction. $f_{i}^{1}$ $f_{i}^{2}$, and $f_{i}^{3}$ denote the representations of DT, while $f_{i}^{'1}$, $f_{i}^{'2}$ and $f_{i}^{'3}$ denote the representations of SS. The similarity between the DT and the SS representations is measured by the cosine distance presented in Eq. (3), 
\begin{equation}
\begin{aligned}
\mathcal S(f, f^{'}) & = 1 - \frac{\langle f,f^{'} \rangle}{|f||f^{'}|},\\
\end{aligned}
\label{eq:equation_03}
\end{equation}
where $f$, $f'$ denote feature vectors, and $\langle \cdot \rangle$ denotes the inner product. Similarities are computed at 3 feature levels, and the weighted sum of them constitutes the $\mathcal{L}_{SS}$ to direct the training of the SS network. The representation of $\mathcal{L}_{SS}$ is as,
\begin{equation}
\begin{aligned}
\mathcal{L}_{SS} & = \frac{1}{N_{2}} \sum_{i=1}^{N_{2}}
\lambda_{1} \mathcal{S}(f_{i}^{1}, f_{i}^{'1}) +
\lambda_{2} \mathcal{S}(f_{i}^{2}, f_{i}^{'2}) +
\lambda_{3} \mathcal{S}(f_{i}^{3}, f_{i}^{'3}).
\\
\end{aligned}
\label{eq:equation_04}
\end{equation}
We denote the parameters of DT and SS as $\theta_{DT}$ and $\theta_{SS}$. The optimization problem for SS is represented as, 
\begin{equation}
\label{eq:equation_05}
\mathop{\arg\min}_{\theta_{SS}} \ \ \ 
E_{z \sim D_{tgt}} 
\mathcal{L}_{SS}(z|\theta_{DT}, \theta_{SS}).
\end{equation}
\begin{algorithm}[t]
\caption{Training of the Stubborn Student Stream}
\begin{algorithmic}[1]
\label{algo_02}
\REQUIRE Target domain training data $D_{tgt}$, DT parameters $\theta_{DT}$, SS density $s\%$, learning rate $\alpha_{2}$, maximum training iterations $K_{2}$, batch size $N_{2}$, regrowth period $T$, and initial regrowth rate $r\%$.

\ENSURE SS parameters $\theta_{SS}$

\STATE Initialize the SS parameters as $\theta_{SS}$.

\STATE Calculate the number of parameters in each convolutional layer. $l_{m}$ denotes the number for the $m$-th layer. 
\STATE Determine the indices $A_{m}$ for the $l_{m} \cdot s\%$ active parameters as in Eq. (6) and $B_{m}$ for the $l_{m} \cdot (1-s\%)$ inactive parameters as in Eq. (7).
\STATE Deactivate the parameters $\omega_{m, n}$ for $n \in B_{m}$ by setting them 0.
\FOR{$k=1$ \TO $K_2$}
\STATE Sample $N_2$ samples $z_{i}$ from $D_{tgt}$.
\STATE Extract features $f_{i}^{1}$, $f_{i}^{2}$, $f_{i}^{3}$ by encoding $z_{i}$ with $\theta_{DT}$, and $f_{i}^{'1}$, $f_{i}^{'2}$, $f_{i}^{'3}$ by encoding $z_{i}$ with $\theta_{SS}$.
\STATE Compute $\mathcal{L}_{SS}$ with $f_{i}^{1}$, $f_{i}^{2}$, $f_{i}^{3}$, $f_{i}^{'1}$, $f_{i}^{'2}$, $f_{i}^{'3}$ as Eq. (4).
\STATE Update the active parameters of $\theta_{SS}$ with $\mathcal{L}_{SS}$, $\alpha_{2}$, $A_{m}$.
\STATE Adjust the regrowth rate $r\%$ with cosine decay.

\IF{$k \  mod \ T \  = \  0$}
\STATE Calculate the number of the active parameters $|A_{m}|$.
\STATE Determine the indices $P_{m}$ for $|A_{m}|\cdot r\%$ parameters to be pruned according to Eq. (8).
\STATE Update the indices of the active and inactive parameters by $A_{m} \gets A_{m} - P_{m}$ and $B_{m} \gets B_{m} \cup P_{m}$.
\STATE Determine the indices $G_{m}$ for $|A_{m}|\cdot r\%$ parameters to be grown according to Eq. (9) - Eq. (12).
\STATE Update the indices of the active and inactive parameters by $A_{m} \gets A_{m} \cup G_{m}$ and $B_{m} \gets B_{m} - G_{m}$.
\ENDIF

\ENDFOR
\RETURN $\theta_{SS}$
\end{algorithmic}
\end{algorithm}
The use of the SS networks multiplies the number of parameters in the face PAD model. The problem is more critical under the client-specific one-class domain adaptation setting since a set of SS networks needs to be stored together with the DT network. To relieve the pressure about the expansion of the model size, sparse SS networks are trained under our framework. The architecture of the SS network is similar to the DT network, and the only difference is that the convolution kernels of the SS network are sparse. For each convolution layer, only $s\%$ number of parameters are non-zero. 

A sparse training strategy is adopted during the optimization of the SS networks. We denote $s\%$ as the desired density of the SS network, and $(1- s\%)$ number of parameters are set as zero and inactive for the update. We use $v_{m, n}$ to denote the initial indicator of the $n$-th parameter in the $m${-th} convolution layer and $\tau_{m}$ to denote the threshold. The inactive parameters are determined by the magnitude of the indicators. The sets of the active and inactive parameter indices for the $m$-th convolution layer are presented in Eq. (\ref{eq:equation_06}) and Eq. (\ref{eq:equation_07}), respectively.
\begin{equation}
\begin{aligned}
A_{m}=\{n| | v_{m, n}| \geq \tau_{m} \},\\
\end{aligned}
\label{eq:equation_06}
\end{equation}
\begin{equation}
\begin{aligned}
B_{m}=\{n| |v_{m, n}| < \tau_{m} \}.\\
\end{aligned}
\label{eq:equation_07}
\end{equation}

During the training, the sets of the active and inactive parameters are periodically adjusted with a regrowth mechanism \cite{SNFS_Axiv19} to optimize the SS network training. The parameter regrowth mechanism consists of the pruning and growing operations. The pruning refers to that the active parameter is set inactive, and the growing refers to the inactive parameter is set active. The operations are applied to the convolution layer of the SS network with a period $T$. After $T$ iterations, the $r\%$ active parameters of each convolution layer will be pruned. The pruning scheme is based on the relative importance of the parameter in the layer, and the importance is measured by the magnitude of the parameter value $|\omega_{m, n}|$. The parameter will be pruned if its magnitude is smaller than the pruning threshold $\tau_{m}^{p}$, and the indices set of parameters to be pruned is represented as 
\begin{equation}
\begin{aligned}
P_{m}=\{n| n \in A_{m} , |\omega_{m, n}| < \tau_{m}^{p} \}.\\
\end{aligned}
\label{eq:equation_08}
\end{equation}
After the pruning, the same number of the inactive parameters will be grown according to their ability to reduce the loss $\mathcal{L}_{SS}$. Following \cite{SNFS_Axiv19}, the estimation of the ability $\mu_{m,n}$ is based on the momentum values $p_{m,n}$ and $q_{m,n}$ as represented in Eq.(\ref{eq:equation_09}) and Eq.(\ref{eq:equation_10}).
\begin{equation}
\begin{aligned}
p_{m,n}^{} & = \beta_{1} p_{m,n}^{'} + (1 -\beta_{1}) \frac{\partial \mathcal{L}_{SS}^{}}{\partial \omega_{m,n}^{}},\\
\end{aligned}
\label{eq:equation_09}
\end{equation}
\begin{equation}
\begin{aligned}
q_{m,n} & = \beta_{2} q_{m,n}^{'} + (1 -\beta_{2}) (\frac{\partial \mathcal{L}_{SS}^{}}{\partial \omega_{m,n}^{}})^2,\\
\end{aligned}
\label{eq:equation_10}
\end{equation}
The $p_{m,n}$ and $q_{m,n}$ are the first and the second order momentum of the current iteration. $p'_{m,n}$ and $q'_{m,n}$ are the first and the second order momentum of the last iteration. $\beta_{1}$ and $\beta_{2}$ are smoothing factors. The value of the momentum $p_{m,n}$ and $q_{m,n}$ can be easily computed with the optimizer class implemented in PyTorch. The computation of the $\mu_{m, n}$ is represented as
\begin{equation}
\begin{aligned}
\mu_{m,n} & = \frac{p_{m,n}}{\sqrt{q_{m,n}}+\epsilon}.\\
\end{aligned}
\label{eq:equation_11}
\end{equation}
$\epsilon$ is a small constant $10^{-8}$ to avoid dividing 0. By comparing the $|\mu_{m, n}|$ with the threshold $\tau_{m}^{g}$, the indices set of parameters to be grown is represented as,
\begin{equation}
\begin{aligned}
G_{m}=\{n| n \in B_{m} , |\mu_{m, n}| \geq \tau_{m}^{g} \}.\\
\end{aligned}
\label{eq:equation_12}
\end{equation}
Fig.\ref{fig:figure_04} illustrates the framework of the SS stream. The Algorithm 2 describes the details about the training of the SS network.

\subsection{Inference at the Test Phase \label{section: method_inference}}
After the student network training, both the teacher network DT and student networks SS are used to compose the inference model. The illustrations for general and client-specific tasks are shown in Fig. \ref{fig:figure_01} and Fig. \ref{fig:figure_02}, respectively. The image sample $I_{t}$ is fed into both the DT and SS networks to generate multi-level feature representations $f_{t}^{1}$, $f_{t}^{2}$, $f_{t}^{3}$, and $f_{t}^{'1}$, $f_{t}^{'2}$, $f_{t}^{'3}$, respectively. The inference score $\xi_{t}$ is computed with the three similarities between features of the two networks, which is represented as,
\begin{equation}
\begin{aligned}
\xi_{t} & = 
\frac{1}{3} ( \mathcal{S}(f_{t}^{1}, f_{t}^{'1}) +
\mathcal{S}(f_{t}^{2}, f_{t}^{'2}) +
\mathcal{S}(f_{t}^{3}, f_{t}^{'3})).\\
\end{aligned}
\label{eq:equation_13}
\end{equation}
As shown in Eq. (\ref{eq:equation_14}), if the score $\xi_{t}$ is smaller than the threshold $\delta$, the image $I_{t}$ is determined as a genuine face. Otherwise, the image $I_{t}$ is determined as an attack.
\begin{equation}
I_{t} =\left\{
\begin{aligned}
genuine, &     & \xi_{t} < \delta,\\
attack, &     &  \xi_{t} \geq \delta.\\
\end{aligned}
\right.
\label{eq:equation_14}
\end{equation}

\section{Experimental Results\label{section: experimental_results}}
To verify the effectiveness of our proposed method, we devise two protocols for the performance evaluation of face PAD models under the general and client-specific one-class domain adaptation settings. We conduct experiments on 5 datasets commonly used for cross-domain face PAD evaluation, including CASIA-FASD \cite{CASIA}, MSU-MFSD\cite{MSU}, IDIAP REPLAY-ATTACK\cite{IDIAP}, NTU ROSE-YOUTU\cite{KSA_TIFS18}, and OULU-NPU \cite{OULU} datasets.

\subsection{Dataset Information}
\subsubsection{CASIA-FASD}
The CASIA Face Anti-spoofing Database (CASIA-FASD) \cite{CASIA} contains 600 video clips collected with 50 subjects. It includes 150 genuine face videos and 450 attack videos. Each subject in the database has 12 corresponding video clips. The videos are collected with 3 different camera sensors, and the attack samples are in printed photos, paper masks, and digital screens.

\subsubsection{IDIAP REPLAY-ATTACK}
The IDIAP REPLAY-ATTACK Database \cite{IDIAP} contains 1200 video clips of 50 subjects, including 200 genuine face videos and 1K attack videos.  With each subject, there are 24 videos collected under controlled and adverse illumination conditions. The attack samples are printed photos, digital face images, and face videos displayed on digital screens.

\subsubsection{MSU-MFSD}
The MSU Mobile Face Presentation Attack Database (MSU-MFSD) \cite{MSU} contains 280 video clips of 35 subjects, including 70 genuine face videos and 210 attack videos. There are 8 videos with each subject. The videos are collected with two types of cameras sensors, and the attack samples are in the type of printed photos and videos replayed on iPhone and iPad.

\subsubsection{NTU ROSE-YOUTU}
The NTU ROSE-YOUTU Face Liveness Detection Dataset \cite{KSA_TIFS18} contains 3350 video clips of 20 subjects, including 1K genuine face videos and 2K attack videos. There are 150-200 video clips for each subject with an average duration of 10 seconds. Five mobile phones with cameras of different specifications have been used for video capture. The attack samples are printed photos, face videos on digital screens, and partial paper masks.

\subsubsection{OULU-NPU}
OULU-NPU Database \cite{OULU} contains 4950 video clips of 55 subjects, including 990 genuine face videos and 3960 attack videos. There are 90 video clips for each subject. The videos are collected by the camera sensors of 6 different types of mobile phones under 3 illumination environments. The attack samples are in the type of printed photos and face videos on digital screens.

\subsection{Evaluation Protocols and Metrics}
\subsubsection{General One-Class Domain Adaptation}
To evaluate the performance of face PAD under the general one-class domain adaptation setting, we devise a protocol named CIMN One-Class Domain Adaptation (CIMN-OCDA) with CASIA-FASD (C), IDIAP REPLAY-ATTACK (I), MSU-MFSD (M), and NTU ROSE-YOUTU (N) datasets, which are commonly used for the evaluation of face PAD methods under the unsupervised domain adaptation settings \cite{KSA_TIFS18, ADA_ICB19, UDA_TIFS20}. For the CIMN-OCDA protocol, each dataset is considered as a data domain. To simulate the cross-domain scenarios, each domain could be set as the source domain and paired with the others as the target domains to form 3 one-class domain adaptation tasks. We denote the tasks by the abbreviations of the datasets. For example, C-I denotes the task with the CASIA-FASD as the source domain and the IDIAP REPLAY-ATTACK as the target domain. For each task, all data of the source domain $train$ set and the genuine face data of the target domain $train$ set are used for model training, while all the data of the target domain $test$ set are used for performance evaluation. In addition to the cross-dataset experiments on the CIMN-OCDA protocol, we also conduct experiments on the OULU One-Class Adaptation Structure A (OULU-OCA-SA) protocol proposed in \cite{OCA-FAS_NC20} for the comparison with similar methods.

\subsubsection{Client-Specific One-Class Domain Adaptation}
For the client-specific one-class domain adaptation task, the target domain is client-specific, and the objective is to develop a specific model with better performance for each target client with a few genuine face images. To evaluate the performance under the client-specific one-class domain adaptation setting, we devise a protocol named CIM-N Client-Specific One-Class Domain Adaptation (CIM-N-CS-OCDA). Since the performance evaluation requires a large amount of data for each target client, we employ the NTU ROSE-YOUTU dataset and sample 10 clients to simulate the client-specific target domains. Each target client domain contains 50 genuine face videos and 110 attack videos. We uniformly divide 25 genuine face videos together with all of 110 attack videos as testing data. Then we sample 1 frame from each of the remaining 25 genuine face videos as the target domain training data. We set the CASIA-FASD (C), IDIAP-REPLAY ATTACK (I), and MSU-MFSD (M) datasets as the source domains to form 3 sub-protocols. For short, we denote the sub-protocols as C-N-CS, I-N-CS, and M-N-CS in the following content. Each sub-protocol contains 10 client-specific one-class domain adaptation tasks for performance evaluation.

\subsubsection{Evaluation Metrics}
The Half Total Error Rate (HTER) and the Area Under Receiver Operating Characteristic Curves (AUC) are the most commonly used metrics for evaluating face PAD methods under cross-domain settings. The HTER is the average of the False Accept Rate (FAR) and the False Reject Rate (FRR), measuring the error rate of the face PAD model at a fixed threshold. As a complement, the AUC is a comprehensive metric that measures the overall performance over different thresholds. Besides, the Average Classification Error Rate (ACER) is the average of the Attack Presentation Classification Error Rate (APCER) and the Bonafide Presentation Classification Error Rate (BPCER). For experiments on the OULU-OCA-SA protocol, we use the ACER and AUC as metrics for a fair comparison with existing methods.

\subsection{Baseline Methods}
To verify the effectiveness of our proposed method, we implement 4 face PAD methods as the baselines under the one-class domain adaptation settings. Besides, we evaluate our method on the OULU-OCA-SA protocol and compare the performance with existing methods addressing face PAD under similar scenarios. Even though our method does not use any target domain attack samples for model training, we compare our method with unsupervised domain adaptation methods, which use unlabelled genuine face and attack samples for model training.

\subsubsection{DT}
As introduced in Section \ref{section: method_DTS}, the depth regression network \cite{BA_CVPR18} is employed as the DT's backbone of the proposed method and trained with the source domain data. Following \cite{BA_CVPR18}, we compute the average value of the pixel map $d_{i}$ as the inference score and set the result as the baseline performance of face PAD without using target domain data for adaptation.

\subsubsection{DT + Fine-Tune}
This method is a naive extension to the DT. The only difference is that we use the genuine face samples of the target domain $train$ set to fine-tune the model after pre-training the DT with the source domain data.

\subsubsection{DT + OCSVM}
Recently, deep neural networks pre-trained on the image classification datasets have been used as feature extractors to develop face PAD models with only some genuine face samples collected in the target domain. Following \cite{CSAD_PR21}, we implement a baseline based on One-Class Support Vector Machine (OCSVM) \cite{SVDD}. We use the pretrained DT with the source domain data as the feature extractor and train an OCSVM classifier for face PAD with the target domain genuine face samples.

\subsubsection{DT + GMM}
Gaussian Mixture Model (GMM) \cite{GMM} is a parametric probability density function, which is commonly used for one-class classification problems. Following \cite{CSAD_PR21}, we also implement a method that trains a GMM-based face PAD model with the DT features as our baseline.  

\subsubsection{OCA-FAS}
OCA-FAS \cite{OCA-FAS_NC20} is a recent method for face PAD under the one-class domain adaptation setting. Since it is the most relevant work to ours in the literature, we also conduct experiments on the OULU-OCA-SA protocol to compare with it.

\subsubsection{Others}
In addition to face PAD methods with one-class domain adaptation, we also compare our method with face PAD methods using unsupervised domain adaptation such as KSA\cite{KSA_TIFS18}, ADA\cite{ADA_ICB19}, UDA\cite{UDA_TIFS20}, USDAN-Un\cite{USDAN-Un_PR21}, etc.

\subsection{Implementation Details}
We apply the same scheme for data pre-processing on all the 5 datasets as follows. We uniformly sample 50 image frames from each video clip. Then the dlib library is used for face detection and alignment. After that, the cropped face regions are further resized to $128 \times 128$. After the data pre-processing, we get 30K, 58.4K, 13.7K, 161.95K, and 246.75K images for CASIA-FASD, IDIAP REPLAY-ATTACK, MSU-MFSD, NTU ROSE-YOUTU, and OULU-NPU datasets. The training of the DT and the SS networks are both optimized with the Adam optimizer\cite{ADAM_ICLR15}. For the training of the DT network, the mini-batch size and learning rate are set as $30$ and $10^{-4}$, respectively. We train the DT network for $8400$ iterations. For the training of the SS network, the mini-batch size and learning rate are set as $25$ and $10^{-4}$, respectively. The weights $\lambda_{1}$, $\lambda_{2}$, $\lambda_{3}$ in Eq. (\ref{eq:equation_04}) are all set as $0.33$. The regrowth period $T$ is set as $60$. The initial regrowth rate $r\%$ is set as $50\%$ and $20\%$ for the experiments of $10\%$ and $1\%$ SS density, and adjusted with cosine decay. We train the SS network for $1500$ iterations. The proposed and baseline methods are implemented based on PyTorch version 1.7.0.

\subsection{General One-Class Domain Adaptation Experiments}
\subsubsection{Experiments on the CIMN-OCDA Protocol}
To verify the effectiveness of our proposed method under the general one-class domain adaptation setting, we firstly conduct experiments on the CIMN-OCDA protocol. For fair comparisons, all the baseline methods and our method share the same DT model as the feature extractor. The SS model density is set as 10\%. We evaluate the cross-dataset performance of face PAD methods under two different experimental settings, which are referred to as the ideal and challenging experimental settings, respectively. For the ideal experimental setting, the HTER performance is directly computed on the test set of the target dataset at the optimal threshold. The experimental results of the ideal setting are shown in Table \ref{tab:table_01}. For the challenging experimental setting, the HTER performance is computed at the threshold that is pre-determined on a validation set where the FRR=10\%. For experiments that use the dataset I as the target dataset, we use the genuine face samples of the development set as the validation set. Since there is no development set or validation set in datasets C, M, and N, for experiments that use C, M, or N as the target dataset, we divide 20\% genuine face samples from the train set of the target dataset as the validation set. The experimental results of the challenging setting are shown in Table \ref{tab:table_02}. Compared to the DT method, our proposed method generally reduces the HTER on different tasks by a clear margin. The reduction of average HTER is more than 10\%, which validates that our method effectively improves the performance of the face PAD model by using only some genuine face samples in the target domain. Moreover, our method outperforms baseline methods with one-class domain adaptation and achieves the best overall performance under both the ideal and challenging experimental settings.
\begin{table*}[htbp]
\centering
\caption{Performance Comparison with the One-Class Domain Adaptation Methods on the CIMN-OCDA Protocol \\ (ideal experimental setting)}
\resizebox{0.9\textwidth}{!}{
\begin{tabular}{|l|c|c|c|c|c|c|c|c|c|c|c|c|c|c|}
\hline
\multirow{2}{*}{Method} & \multicolumn{13}{c|}{HTER (\%) ↓ }\\
\cline{2-14}
& C-I 
& C-M 
& C-N 
& I-C 
& I-M 
& I-N 
& M-C 
& M-I 
& M-N 
& N-C 
& N-I 
& N-M 
& Average\\
\hline
\hline
DT & 36.3 & \textbf{14.1} & 28.4 & 45.1 & 35.9 & 43.2 & 35.0 & 23.4 & 38.2 & 28.9 & 29.0 & 21.7 & 31.6\\
\hline
DT + Fine-tune & 41.1 & 23.2 & 35.7 & 49.4 & \textbf{18.6} & 35.7 & 29.5 & 36.0 & 36.2 & \textbf{19.7} & 27.7 & 19.4 & 31.0\\
\hline
DT + OCSVM \cite{CSAD_PR21} & 16.2 & 25.9 & 31.4 & 34.8 & 19.0 & 37.8 & 39.7 & 8.4 & 30.4 & 26.2 & 18.1 & 25.4 & 26.1 \\
\hline
DT + GMM \cite{CSAD_PR21} & 8.6 & 28.2 & 34.8 & \textbf{23.8} & 19.2 & \textbf{30.8} & 27.6 & 6.3 & 27.8 & 22.6 & 11.4 & 21.0 & 21.8\\
\hline
\hline
Ours & \textbf{3.5} & 15.0 & \textbf{26.9} & 31.9 & 20.8 & 31.1 & \textbf{26.7} & \textbf{2.9} & \textbf{27.2} & 21.7 & \textbf{3.0} & \textbf{10.6} & \textbf{18.4}\\
\hline
\end{tabular}
}
\label{tab:table_01}
\end{table*}

\begin{table*}[htbp]
\centering
\caption{Performance Comparison with the One-Class Domain Adaptation Methods on the CIMN-OCDA Protocol \\ (challenging experimental setting with pre-determined threshold on the validation set where FRR=10\%)}
\resizebox{0.9\textwidth}{!}{
\begin{tabular}{|l|c|c|c|c|c|c|c|c|c|c|c|c|c|c|}
\hline
\multirow{2}{*}{Method} & \multicolumn{13}{c|}{HTER (\%) ↓ }\\
\cline{2-14}
& C-I 
& C-M 
& C-N 
& I-C 
& I-M 
& I-N 
& M-C 
& M-I 
& M-N 
& N-C 
& N-I 
& N-M 
& Average\\
\hline
DT & 54.8 & 20.6 & 28.8 & 58.2 & 40.0 & 49.0 & 42.0 & 46.3 & 41.1 & 39.3 & 55.2 & 25.3 & 41.7\\
\hline
DT + Fine-tune & 44.0 & 22.1 & 35.2 & 41.6 & 23.7 & 39.3 & 29.7 & 36.3 & 33.0 & 31.0 & 29.5 & 35.2 & 33.4\\
\hline
DT + OCSVM [25] & 23.0 & 26.0 & 33.3 & \textbf{33.4} & 27.9 & 40.8 & 41.8 & 9.2 & 30.6 & 25.7 & 23.2 & 28.8 & 28.6\\
\hline
DT + GMM [25] & 11.7 & 33.6 & 34.7 & 34.7 & \textbf{20.3} & 32.1 & 32.7 & 6.9 & 27.5 & 22.2 & 13.5 & 23.8 & 24.5\\
\hline
\hline
Ours & \textbf{3.8} & \textbf{17.6} & \textbf{28.5} & 33.7 & 28.1 & \textbf{31.0} & \textbf{29.0} & \textbf{4.3} & \textbf{27.0} & \textbf{21.4} & \textbf{3.0} & \textbf{20.4} & \textbf{20.7}\\
\hline
\end{tabular}
}
\label{tab:table_02}
\end{table*}
\begin{table*}[tbp]
\centering
\caption{Performance Comparison With the State-of-the-art Unsupervised Domain Adaptation Methods}
\resizebox{0.9\textwidth}{!}{
\begin{tabular}{|l|c|c|c|c|c|c|c|c|c|c|c|c|c|c|}
\hline
\multirow{2}{*}{Method} & \multicolumn{13}{c|}{HTER (\%) ↓ }\\
\cline{2-14}
& C-I 
& C-M 
& C-N 
& I-C 
& I-M 
& I-N 
& M-C 
& M-I 
& M-N 
& N-C 
& N-I 
& N-M 
& Average\\
\hline
\hline
ADDA \cite{ADDA_CVPR17} & 41.8 & 36.6 & 31.4 & 49.8 & 35.1 & 50.0 & 39.0 & 35.2 & 38.7 & 28.7 & 34.6 & 33.4 & 37.9 \\
\hline
DRCN \cite{DRCN_ECCV16} & 44.4 & 27.6 & 32.5 & 48.9 & 42.0 & 50.0 & 28.9 & 36.8 & 39.4 & 32.3 & 37.4 & 37.2 & 38.1 \\
\hline
DupGAN \cite{DUPGAN_CVPR18} & 42.4 & 33.4 & 30.8 & 46.5 & 36.2 & 47.0 & 27.1 & 35.4 & 34.5 & 24.6 & 35.9 & 33.4 & 35.6 \\
\hline
USDAN-Un \cite{USDAN-Un_PR21} & 16.0 & 9.2 & / & 30.2 & 25.8 & / & 13.3 & 3.4 & / & / & / & / & / \\
\hline
KSA \cite{KSA_TIFS18} & 39.3 & 15.1 & 31.6 & \textbf{12.3} & 34.9 & 40.1 & \textbf{9.1} & 33.3 & 30.4 & 30.1 & 38.8 & 26.1 & 28.4 \\
\hline
ADA \cite{ADA_ICB19} & 17.5 & 9.3 & 29.4 & 41.5 & 30.5 & 41.7 & 17.7 & 5.1 & 32.7 & 34.1 & 30.3 & 31.5 & 26.8 \\
\hline
ML-Net \cite{UDA_TIFS20} & 43.3 & 14.0 & 32.4 & 45.4 & 35.3 & 42.8 & 37.8 & 11.5 & 34.6 & 25.7 & 30.7 & 32.6 & 32.2\\
\hline
UDA \cite{UDA_TIFS20} & 15.6 & \textbf{9.0} & 28.0 & 34.2 & 29.0 & 39.8 & 16.8 & 3.0 & 29.7 & \textbf{17.9} & 23.7 & 24.4 & 22.6 \\
\hline
\hline
Ours & \textbf{3.5} & 15.0 & \textbf{26.9} & 31.9 & \textbf{20.8} & \textbf{31.1} & 26.7 & \textbf{2.9} & \textbf{27.2} & 21.7 & \textbf{3.0} & \textbf{10.6} & \textbf{18.4}\\
\hline
\end{tabular}
}
\label{tab:table_03}
\end{table*}

\begin{table}[tbp]
\centering
\caption{Performance Comparison with Baseline Methods on the OULU-OCA-SA Protocol}
\resizebox{0.35\textwidth}{!}{
\begin{tabular}{|l|c|c|}
\hline
Method & ACER (\%) ↓ & AUC (\%) ↑\\
\hline
\hline
DTN \cite{DTN_CVPR19} & 15.61±1.69 & /\\
\hline
OCA-FAS \cite{OCA-FAS_NC20} & 2.26±0.39 & /\\
\hline
DT & 3.95±0.30 & 98.17±0.17\\
\hline
Ours & \textbf{0.46±0.12} & \textbf{99.63±0.12}\\
\hline
\end{tabular}
}

\label{tab:table_04}
\end{table}

In addition, we also compare our method with state-of-the-art face PAD methods with unsupervised domain adaptation. We find that our proposed method achieves better performance with less target domain data for model training as shown in Table~\ref{tab:table_03}. Our proposed method achieves the best performance on 8 out of 12 experiments and outperforms the state-of-the-art methods by 4.2\% in terms of average HTER. The advantage of our proposed method is especially significant on C-I, I-M, I-N, N-I, and N-M experiments, where our method reduces the HTER by more than 12.1\%, 5.0\%, 8.7\%, 20.7\%, and 13.8\% compared to the state-of-the-art.

\subsubsection{Experiments on the OULU-OCA-SA Protocol}
To compare with OCA-FAS \cite{OCA-FAS_NC20}, which is a recent method for the face PAD with one-class domain adaptation, we evaluate our proposed method on the OULU-OCA-SA protocol. We conduct experiments on Protocol 3, the most challenging task on OULU-OCA-SA, and the SS model density of our method is set as 10\%. From the experimental results shown in Table~\ref{tab:table_04}, we find that our proposed method outperforms the DTN and OCA-FAS by a clear margin. The ACER and the AUC of our proposed method are $0.46\%$ and $99.63\%$, respectively. Compared to our baseline method, using the target domain genuine face images for domain adaptation helps to reduce the ACER by $3.49\%$ and improve the AUC by $1.46\%$.

\begin{table*}[htbp]
\centering
\caption{Performance Comparison with the One-class Domain Adaptation Methods on the CIM-N-CS-OCDA Protocol}
\resizebox{0.9\textwidth}{!}{
\begin{tabular}{|c|l|c|c|c|c|c|c|c|c|c|c|c|c|}
\hline
\multirow{2}{*}{Protocol} & \multirow{2}{*}{Methods} & \multicolumn{11}{c|}{HTER (\%) ↓ }\\
\cline{3-13}
& 
& Client 1 & Client 2 & Client 3 & Client 4 & Client 5
& Client 6 & Client 7 & Client 8 & Client 9 & Client 10
& Overall\\
\hline
\hline
\multirow{6}{*}{I-N-CS}
& DT
& 36.16	& 47.42	& 30.72	& 32.71 & 25.49	
& 47.00	& 47.95	& 43.18	& 41.60 & 46.51
& 39.87±8.07 \\

\cline{2-13} 
& DT + Fine-tune
& 22.99	& 22.52	& 22.57	& 16.97	& 23.35	
& 13.15	& 35.25	& 22.42	& 32.42	& 24.50	
& 23.61±6.43 \\

\cline{2-13}
& DT + OCSVM \cite{CSAD_PR21}
& 26.55	& 21.83	& 15.53	& 9.19	& 16.20	
& 20.07	& 24.41	& 13.80	& 16.11	& 29.49	
& 19.32±6.29 \\

\cline{2-13}
& DT + GMM \cite{CSAD_PR21}
& 24.13	& 20.38	& 18.72	& 10.96	& 17.56	
& 15.87	& 26.45	& 20.71	& 16.12	& 25.37	
& 19.63±4.81 \\

\cline{2-13}
& Ours
& 18.97	& 18.20	& 13.05	& 7.89	& 16.21	
& 11.53	& 20.13	& 14.13	& 10.22	& 16.24	
& \textbf{14.66±4.00} \\

\hline
\hline
\multirow{6}{*}{M-N-CS}
& DT
& 43.13	& 38.57	& 38.49	& 42.14	& 37.05	
& 39.49	& 29.79	& 30.93	& 40.25	& 32.93	
& 37.28±4.60 \\

\cline{2-13} 
& DT + Fine-tune
& 22.32	& 19.09	& 14.54	& 11.04	& 25.42	
& 25.55	& 14.49	& 12.55	& 12.97	& 19.64	
& 17.76±5±39 \\

\cline{2-13}
& DT + OCSVM \cite{CSAD_PR21}
& 21.61	& 18.47	& 11.68	& 13.79	& 13.38	
& 24.56	& 17.17	& 13.66	& 11.41	& 19.81	
& 16.55±4.48 \\

\cline{2-13}
& DT + GMM \cite{CSAD_PR21}
& 21.50	& 17.63	& 11.74	& 11.64	& 15.57	
& 19.61	& 21.36	& 14.99	& 11.53	& 19.92	
& 16.55±4.02 \\

\cline{2-13}
& Ours
& 12.34	& 18.06	& 7.92	& 8.13	& 13.41	
& 14.23	& 13.84	& 11.01	& 10.98	& 16.54	
& \textbf{12.65±3.29} \\

\hline
\hline
\multirow{6}{*}{C-N-CS}
& DT
& 31.39	& 28.11	& 21.68	& 32.88	& 27.54	
& 26.48	& 29.60	& 30.35	& 26.14	& 28.53	
& 28.27±3.14 \\

\cline{2-13} 
& DT + Fine-tune
& 23.83	& 23.48	& 16.19	& 11.27	& 24.59	
& 18.73	& 10.86	& 15.37	& 19.39	& 17.50	
& 18.12±4.90 \\

\cline{2-13}
& DT + OCSVM \cite{CSAD_PR21}
& 23.17	& 23.03	& 11.39	& 7.40	& 17.67	
& 21.87	& 13.25	& 17.80	& 12.48	& 25.19	
& 17.33±5.99 \\

\cline{2-13}
& DT + GMM \cite{CSAD_PR21}
& 21.04	& 20.12	& 7.16	& 6.51	& 20.24	
& 16.20	& 22.74	& 19.00	& 13.82	& 21.71	
& 16.85±5.90 \\

\cline{2-13}
& Ours
& 6.22	& 9.89	& 3.75	& 6.07	& 8.05	
& 10.19	& 14.94	& 6.43	& 10.96	& 9.14	
& \textbf{8.56±3.18} \\
\hline
\end{tabular}
}

\label{tab:table_05}
\end{table*}

\begin{table*}[htbp]
\centering
\caption{Performance Comparison with the One-class Domain Adaptation Methods on the CIM-N-CS-OCDA Protocol}
\resizebox{0.9\textwidth}{!}{
\begin{tabular}{|c|l|c|c|c|c|c|c|c|c|c|c|c|c|}
\hline
\multirow{2}{*}{Protocol} & \multirow{2}{*}{Methods} & \multicolumn{11}{c|}{AUC (\%) ↑ }\\
\cline{3-13}
& 
& Client 1 & Client 2 & Client 3 & Client 4 & Client 5
& Client 6 & Client 7 & Client 8 & Client 9 & Client 10
& Overall\\
\hline
\hline
\multirow{6}{*}{I-N-CS}
& DT
& 64.56 & 44.10 & 70.95 & 72.66 & 79.55	
& 43.58 & 47.95 & 57.65 & 58.00 & 53.14	
& 59.21±12.48\\
\cline{2-13} 
& DT + Fine-tune
& 79.98 & 81.02 & 82.25 & 90.19 & 79.43	 
& 93.66 & 66.37 & 82.05 & 69.36 & 81.16	
& 80.55±8.15\\

\cline{2-13}
& DT + OCSVM \cite{CSAD_PR21}
& 82.05 & 85.51 & 93.19 & 95.56 & 90.15	
& 85.59 & 79.18 & 91.26 & 89.59 & 77.24	
& 86.93±6.07\\
\cline{2-13}
& DT + GMM \cite{CSAD_PR21}
& 81.30 & 85.98 & 89.72 & 94.29 & 87.81	
& 88.30 & 79.97 & 86.70 & 88.22 & 81.94	
& 86.42±4.34\\
\cline{2-13}
& Ours
& 89.54 & 88.92 & 92.61 & 96.92 & 89.05	
& 92.62 & 86.53 & 93.29 & 95.61 & 92.16	
& \textbf{91.73±3.22}\\
\hline
\hline
\multirow{6}{*}{M-N-CS}
& DT
& 58.01 & 63.94 & 62.60 & 60.13 & 62.98	
& 60.40 & 77.97 & 73.60 & 63.29 & 68.07	
& 65.10±6.32\\
\cline{2-13} 
& DT + Fine-tune
& 81.99 & 84.08 & 90.84 & 93.37 & 79.01	
& 74.73 & 89.66 & 93.83 & 92.48 & 84.16	
& 86.42±6.60\\

\cline{2-13}
& DT + OCSVM \cite{CSAD_PR21}
& 85.80 & 88.63 & 94.51 & 93.09 & 93.93	
& 82.72 & 90.67 & 91.02 & 94.66 & 85.18	
& 90.02±4.27\\
\cline{2-13}
& DT + GMM \cite{CSAD_PR21}
& 85.90 & 89.66 & 94.91 & 95.42 & 90.71	
& 88.22 & 83.70 & 92.56 & 93.29 & 88.65	
& 90.30±3.83\\
\cline{2-13}
& Ours
& 93.83 & 90.64 & 97.33 & 97.70 & 89.98	
& 93.29 & 90.32 & 95.62 & 95.79 & 91.74	
& \textbf{93.62±2.90}\\
\hline
\hline
\multirow{6}{*}{C-N-CS}
& DT
& 74.22 & 77.39 & 86.52 & 73.49 & 77.40	
& 78.65 & 80.52 & 76.77 & 78.66 & 80.63	
& 78.43±3.68\\
\cline{2-13} 
& DT + Fine-tune
& 82.08 & 85.69 & 88.93 & 91.22 & 84.28	
& 85.56 & 89.87 & 90.25 & 82.76 & 88.50	
& 86.91±3.26\\
\cline{2-13}

& DT + OCSVM \cite{CSAD_PR21}
& 82.89 & 83.81 & 95.73 & 96.16 & 90.13	
& 86.38 & 91.41 & 87.03 & 91.53 & 80.66	
& 88.57±5.30\\
\cline{2-13}
& DT + GMM \cite{CSAD_PR21}
& 85.65 & 84.87 & 96.57 & 96.84 & 88.41	
& 91.97 & 80.72 & 86.35 & 91.22 & 86.37	
& 88.90±5.20\\
\cline{2-13}
& Ours
& 97.52 & 95.31 & 99.18 & 98.44 & 96.20	
& 96.12 & 91.66 & 98.42 & 96.72 & 96.70	
& \textbf{96.63±2.13}\\
\hline
\end{tabular}
}

\label{tab:table_06}
\end{table*}

\subsection{Client-Specific One-Class Domain Adaptation Experiments}
In addition to the experiments under the general one-class domain adaptation setting, we also conduct experiments under the client-specific domain adaptation setting to verify that our proposed method can improve the performance of face PAD for the specific target client by only using a few genuine face images for one-class domain adaptation.

\begin{figure}[htbp]
\centering
\subfloat[Score Distribution before One-Class Domain Adaptation]{
\includegraphics[width=0.45\textwidth]{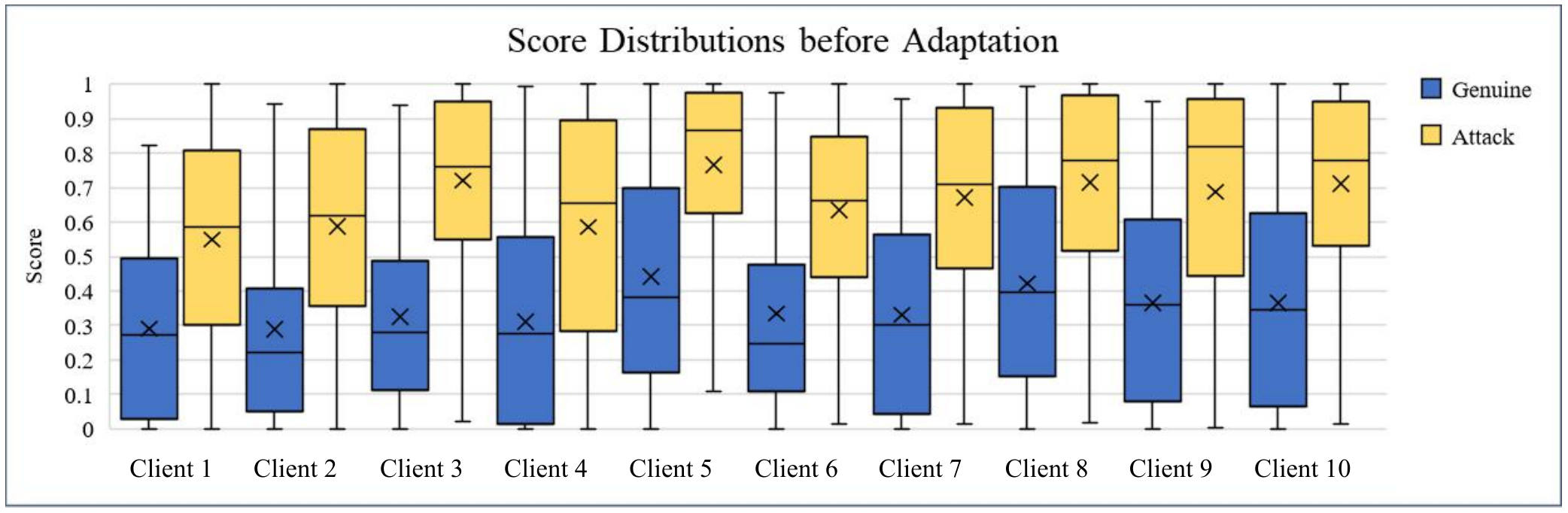}
}
\quad
\subfloat[Score Distribution after One-Class Domain Adaptation]{
\includegraphics[width=0.45\textwidth]{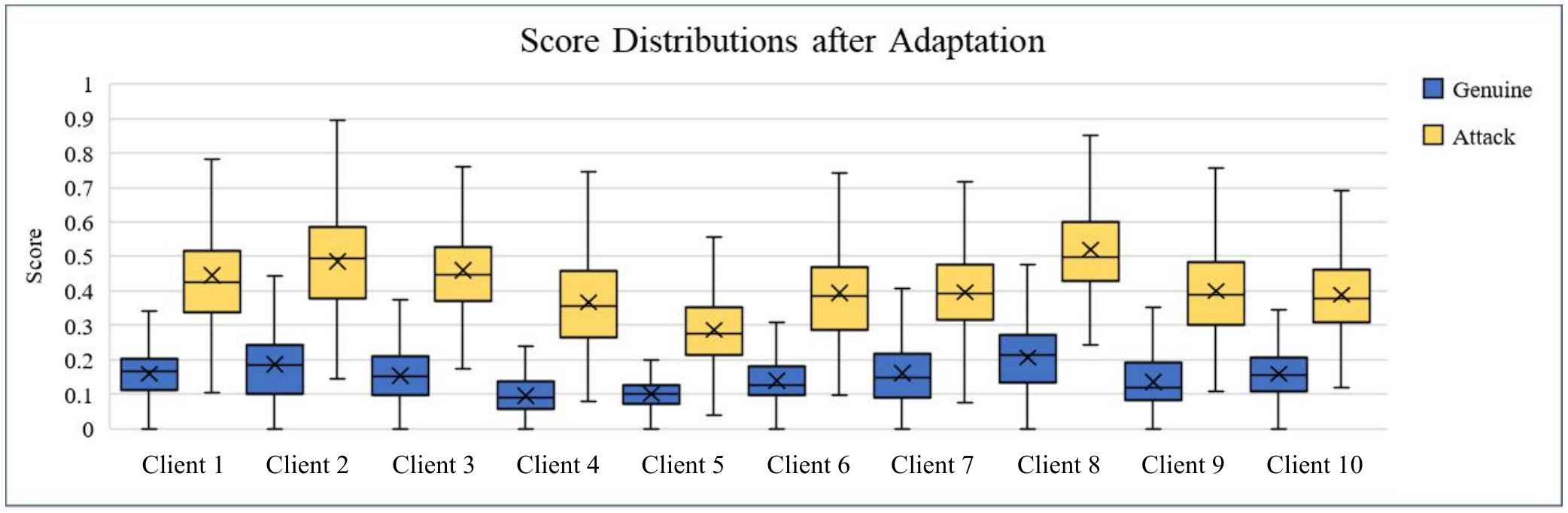}
}
\caption{The figure (a) shows the score distributions of the DT on 10 tasks of C-N-CS protocol. A pair of blue and yellow boxes are used to illustrate the score distribution for each task. The blue and yellow boxes illustrate the distributions of the testing genuine face and attack samples, respectively.}
\label{fig:figure_05}
\end{figure}
The experiments are conducted on the CIM-N-CS-OCDA protocol. The SS model density is set as 10\% and we evaluate the performance with both HTER and AUC. From the experimental results shown in Table~\ref{tab:table_05} and Table~\ref{tab:table_06}, our method significantly improves the face PAD performance and consistently outperforms baseline methods on all the three sub-protocols. Compared to the DT baseline without using any target domain data for adaptation, our method reduces the average HTER by $25.21\%$, $24.63\%$, $19.71\%$ and improves the average AUC by $32.52\%$, $28.52\%$, $18.20\%$ on the I-N-CS, M-N-CS, C-N-CS sub-protocols, respectively. Compared to baseline methods that use target domain genuine face samples for model training, our method also shows distinct advantages. On the I-N-CS sub-protocol, our method outperforms baseline methods by more than $4.66\%$ and $4.80\%$ in HTER and AUC. On the M-N-CS and C-N-CS sub-protocols, the improvement are more than $3.90\%$, $8.29\%$ in terms of HTER and $3.32\%$, $7.73\%$ in terms of AUC. Comparing the performance of our proposed method on different sub-protocols, we find that even though we use the same target domain data to train the SS networks, there are distinct performance differences between models with different DT networks. The results indicate that the discrimination ability of the DT's feature representations will influence the final performance of the proposed method.

To intuitively demonstrate the effectiveness of our method, we visualize the score distributions on the C-N-CS sub-protocol. Fig. \ref{eq:equation_05}-(a) shows the score distributions of the DT baseline. The blue and yellow boxes represent the genuine and attack samples, respectively. The distributions of both classes are wide-spreading, and there are obvious overlaps between the genuine face and attack distributions. Fig. \ref{eq:equation_05}-(b) shows the score distributions of our method. With the help of the genuine face samples in the target domain, the score distributions become more compact and separable between genuine and attack samples. As shown in Fig. \ref{eq:equation_06}, we compare our method with other baseline methods using target domain genuine samples for model training. Although the score distributions of the DT + GMM and DT + OCSVM are as compact as ours, our method has the largest separation gap between the distributions of the genuine face and attack samples.

\begin{figure}[t]
\centering
\includegraphics[width=0.45\textwidth]{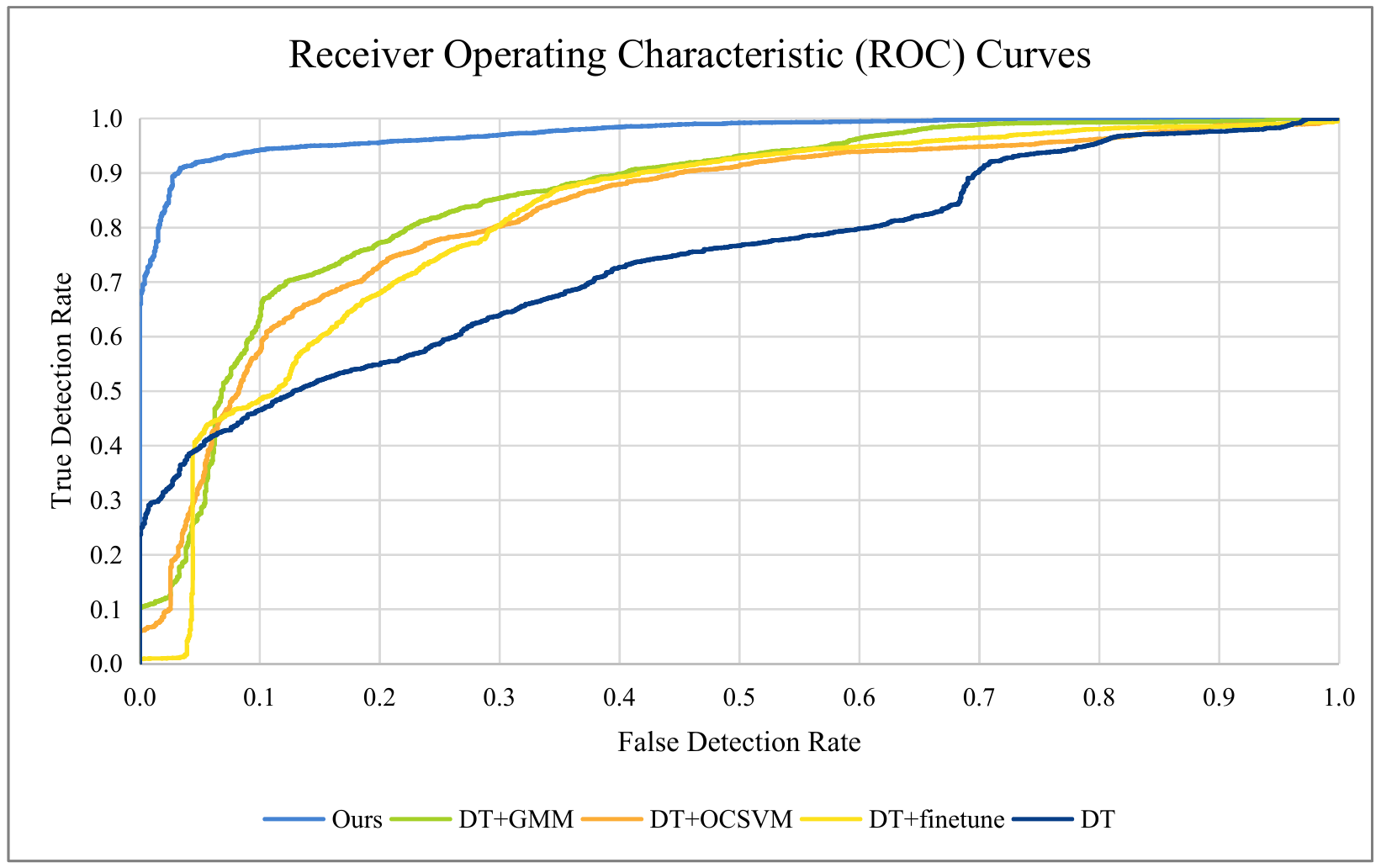}
\caption{The figure shows the Receiver Operating Characteristic (ROC) Curves of different methods on C-N-CS task 1. The horizontal axis is the False Detection Rate (FDR) and the vertical axis is the True Detection Rate (TDR).}
\label{fig:figure_07}
\end{figure}

Besides, we also visualize the Receiver Operating Characteristic Curves of proposed and baseline methods on the C-N-CS task 1 as in Fig. \ref{fig:figure_07}. As shown in the figure, the True Detection Rate (TDR) of our method is higher than the TDRs of baseline methods at different False Detection Rate (FDR) conditions. The advantage of our method is especially greater at the thresholds where the FDR is small.
\begin{figure}[tbp]
\centering
\includegraphics[width=0.45\textwidth]{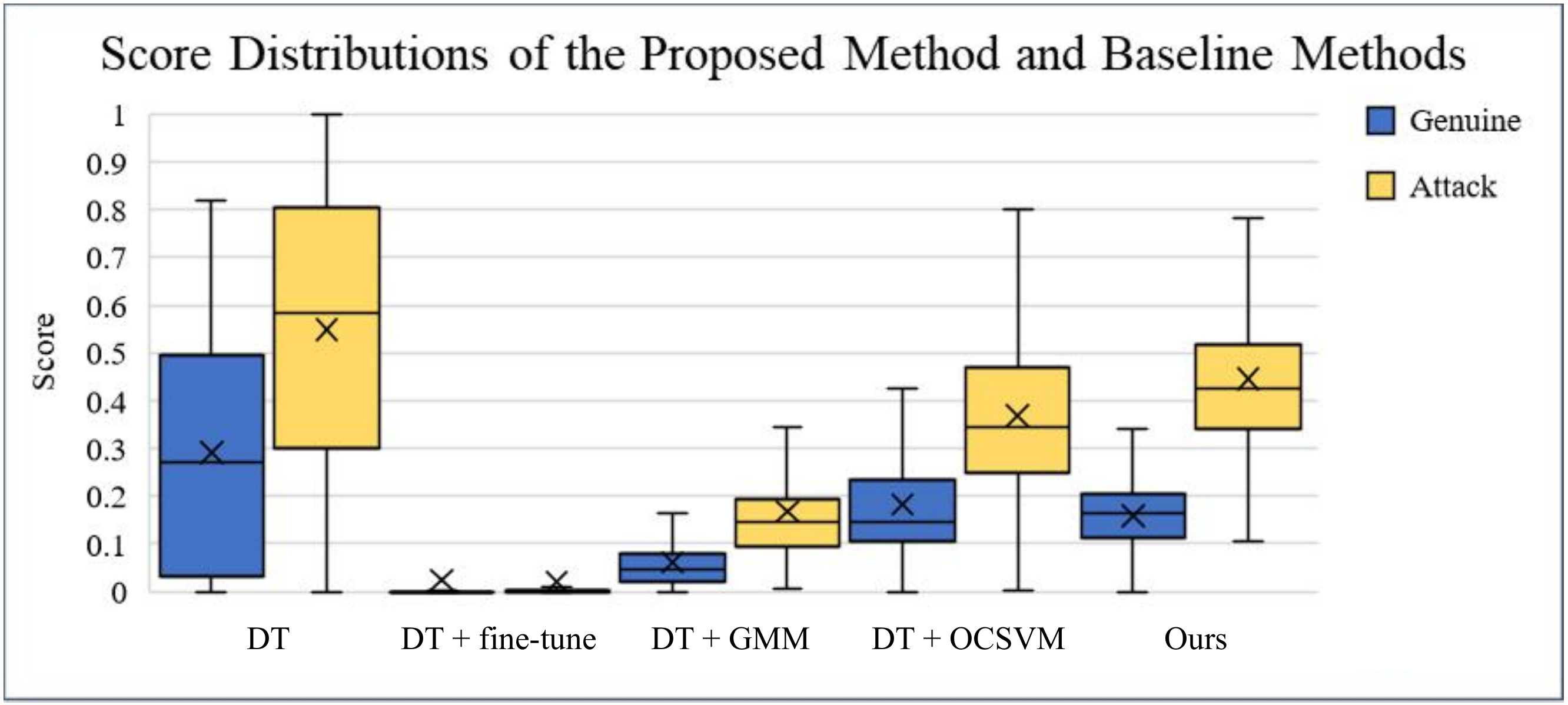}
\caption{The figure shows the score distributions of different methods on C-N-CS task 1. From the left to the right are score distributions of the DT, DT+finetune, DT+GMM, DT+OCSVM and our proposed method. The blue and yellow boxes illustrate the distributions of the genuine face and attack samples, respectively. The figure shows that our method has larger separation gap compared to baseline methods.}
\label{fig:figure_06}
\end{figure}

\begin{table}[t]
\centering
\caption{Performance of the Proposed Method with Different Stubborn Student Density on the CIM-N-CS-OCDA Protocol}
\resizebox{0.45\textwidth}{!}{
\begin{tabular}{|c|l|c|c|}
\hline
Protocol & SS Density (\%) & HTER (\%) ↓ & AUC (\%) ↑\\
\hline
\hline
\multirow{3}{*}{I-N-CS}
& 100 & 15.19 & 91.26 \\
\cline{2-4}
& 10  & \textbf{14.66} & \textbf{91.73}\\
\cline{2-4}
& 1	& 16.94 & 89.26\\
\hline
\hline
\multirow{3}{*}{M-N-CS}
& 100 & 14.62 & 92.60\\
\cline{2-4}
& 10  & \textbf{12.65} & \textbf{93.62}\\
\cline{2-4}
& 1   & 13.16 & 92.85\\
\hline
\hline
\multirow{3}{*}{C-N-CS}
& 100 & 8.65  & 96.50\\
\cline{2-4}
& 10  & \textbf{8.56}  & \textbf{96.63}\\
\cline{2-4}
& 1	& 10.71 & 95.03\\
\hline
\end{tabular}
}
\label{tab:table_07}
\end{table}

\begin{table}[ht]
\centering
\caption{Model Size of the Stubborn Student at Different Density}
\resizebox{0.45\textwidth}{!}{
\begin{tabular}{|l|c|c|}
\hline
SS Density (\%) & No. Non-zero Params (M) & Memory (MB) \\
\hline
\hline
100 & 1.73 & 6.62 \\
\hline
10  & 0.18 & 2.02 \\
\hline
1	  & 0.02 & 0.25\\
\hline
\end{tabular}
}
\label{tab:table_08}
\end{table}
\subsection{Ablation Study}
Besides, we also conduct experiments to analyze the impacts of the SS density, the parameter regrowth mechanism, and the multi-level distillation scheme.

\subsubsection{Impact of the Stubborn Student Density}
To study the impact of the SS Density, we conduct experiments on the CIM-N-CS-OCDA protocols. The HTER and AUC at the SS density level $100\%$, $10\%$, and $1\%$ are evaluated, and the results are shown in Table~\ref{tab:table_07}. On the I-N-CS and C-N-CS sub-protocols, the models with the $10\%$ SS density achieve comparable performance to the models with the $100\%$ SS density and perform better than the models with the $1\%$ SS density. On the M-N-CS sub-protocol, the models with $10\%$ and $1\%$ SS density slightly outperform the models with $100\%$ SS density. The results indicate that the SS of $100\%$ density is over-parametric to learn genuine face representations of the DT network. The moderate sparsity of the SS network helps to reduce the model size without severely losing accuracy. To verify the feasibility of model compression, we measure the model size of the SS at different density levels. As shown in Table~\ref{tab:table_08}, the number of non-zero parameters at $100\%$, $10\%$, and $1\%$ density are $1.73$ M, $0.18$ M, and $0.02$ M. We save the sparse models in Coordinate (COO) data format and the size of the SS model with $100\%$, $10\%$, and $1\%$ density are 6.62 MB, 2.02 MB, and 0.25 MB, respectively. Note that the compression rate of the model size in terms of memory is not the same as the number of non-zero parameters. It is because that we need to store the coordinates for the non-zero parameters of the sparse model.

\subsubsection{Impact of the Parameter Regrowth Mechanism}
\begin{figure}[tbp]
\centering
\includegraphics[width=0.45\textwidth]{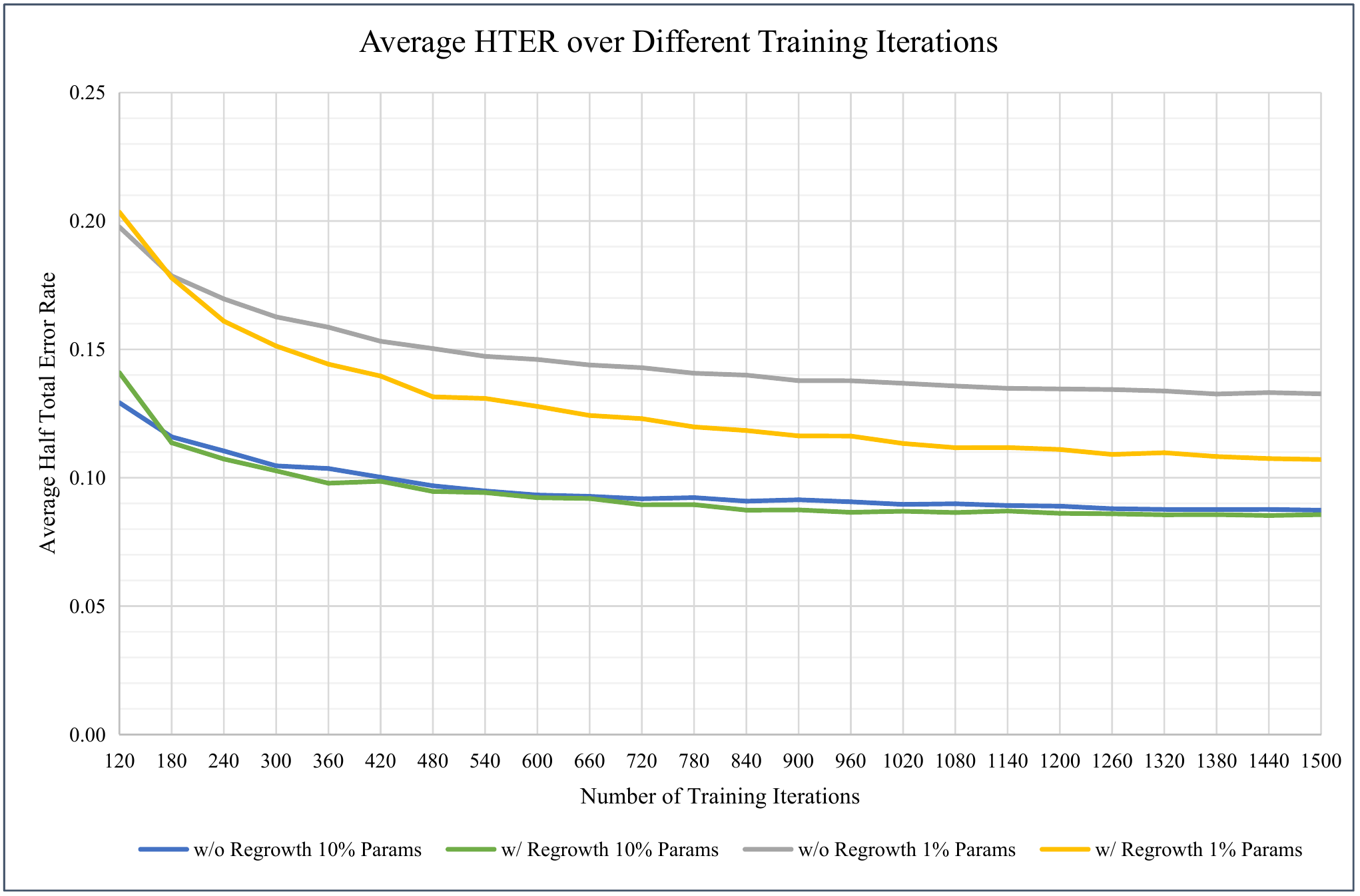}
\caption{This figure shows the average HTER of our proposed method over the number of training iterations on C-N-CS tasks. The four curves illustrate the impact of the regrowth mechanism on the proposed method at two SS model density conditions.}
\label{fig:figure_08}
\end{figure}

To analyze the impact of the parameter regrowth mechanism, we conduct ablation experiments on the C-N-CS sub-protocol and plot the changing of the average HTER with the number of training iterations in Fig. \ref{fig:figure_08}. We can see that the parameter regrowth mechanism helps optimize the training process, especially when the SS network is at a lower density.

\begin{table}[t]
\centering
\caption{Performance of the Proposed Method with Single-Level and Multi-Level Distillation on the CIM-N-CS-OCDA Protocol}
\resizebox{0.45\textwidth}{!}{
\begin{tabular}{|c|l|c|c|}
\hline
Protocol & Distillation Level& HTER (\%) ↓ & AUC (\%) ↑\\
\hline
\hline
\multirow{2}{*}{I-N-CS}
& single-level & \textbf{14.15} & \textbf{92.37} \\
\cline{2-4}
& multi-level  & 14.66  & 91.73\\
\hline
\hline
\multirow{2}{*}{M-N-CS}
& single-level & 14.68 & 92.66 \\
\cline{2-4}
& multi-level  & \textbf{12.65} & \textbf{93.62}\\
\hline
\hline
\multirow{2}{*}{C-N-CS}
& single-level & 10.34 & 95.41 \\
\cline{2-4}
& multi-level  & \textbf{8.56}  & \textbf{96.63}\\
\hline
\end{tabular}
}
\label{tab:table_09}
\end{table}

\subsubsection{Impact of the Multi-Level Distillation}
To analyze the impact of the multi-level distillation scheme, we conduct comparison experiments on the CIM-N-CS-OCDA protocols, and the results are shown in Table~\ref{tab:table_09}. On the I-N-CS sub-protocol, our method with single-level and multi-level distillation achieve comparable results in the HTER, but the single-level distillation one has better AUC performance. On the M-N-CS and C-N-CS sub-protocols, our method with multi-level distillation generally outperforms the single-level one by $2.03\%$, $1.78\%$ in  HTER and $0.96\%$, $1.22\%$ in AUC.

\subsection{Limitations and Future Work}
Although our proposed method generally outperforms the state-of-the-art methods, there is still room for performance improvement under the general and client-specific one-class domain adaptation setting. The HTER performance is not satisfactory in some experimental settings, especially where a simple dataset is used as the source domain and a more complex dataset as the target domain, such as I-C, I-N, and M-N in Table \ref{tab:table_01}.

\begin{figure}[tbp]
\centering
\includegraphics[width=0.45\textwidth]{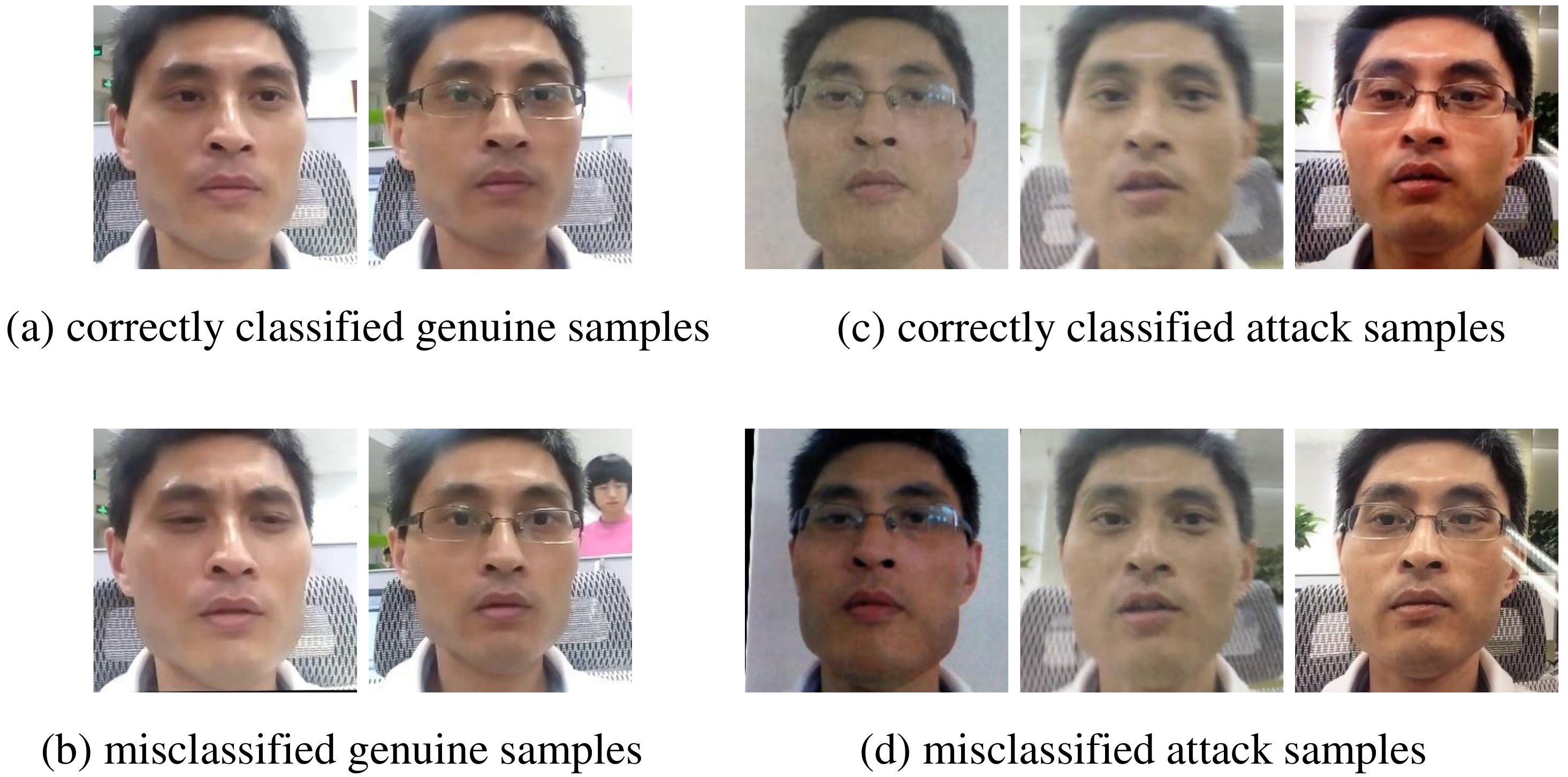}
\caption{The figures show examples of the image samples correctly classified and misclassified by our model on C-N-CS task 1. Figure (a) shows the correctly classified genuine face samples; figure (b) shows the misclassified genuine face samples; figure (c) shows the correctly classified attack samples; figure (d) shows the misclassified attack samples.}
\label{fig:figure_09}
\end{figure}

Besides, the classification accuracy of our model for some difficult samples needs to be further improved. As shown in Fig. \ref{fig:figure_09}, some genuine face samples with unusual facial expressions or backgrounds are misclassified as attacks; some attack samples with natural skin color and higher definition are misclassified as genuine face images.

From the experimental results in Table \ref{tab:table_05}, Table \ref{tab:table_06}, and Table \ref{tab:table_09}, even use the same target domain data to train the student network SS, the performance of the proposed method varies with the feature quality of the teacher network DT. Therefore, a promising direction for future work is to further improve the feature learning and feature selection of the teacher network.

\section{Conclusions}
In this paper, we introduce a framework to address the cross-domain problem in face PAD with one-class domain adaptation, which improves the cross-domain performance of the face PAD model by utilizing only a few genuine face samples collected in the target domain. Under the framework, a teacher network is trained with genuine face and attack samples of the source domain to provide multi-level discriminative feature representations for face PAD. Student networks are trained with only genuine face samples of the target domain to generate similar representations to the teacher's outputs. To verify the effectiveness of our method under one-class domain adaptation settings, we devised two new protocols on public face PAD datasets and did extensive experiments. The experimental results show that our method outperforms baseline methods under one-class domain adaptation settings and even performs better than state-of-the-art methods with unsupervised domain adaptation. However, there is still room for performance improvement in some experimental settings. In future work, we will explore methods to further improve the feature learning and feature selection of the teacher network.

% references section
\bibliographystyle{IEEEtran}
\bibliography{reference.bib}

% Generated by IEEEtran.bst, version: 1.14 (2015/08/26)
\begin{thebibliography}{10}
\providecommand{\url}[1]{#1}
\csname url@samestyle\endcsname
\providecommand{\newblock}{\relax}
\providecommand{\bibinfo}[2]{#2}
\providecommand{\BIBentrySTDinterwordspacing}{\spaceskip=0pt\relax}
\providecommand{\BIBentryALTinterwordstretchfactor}{4}
\providecommand{\BIBentryALTinterwordspacing}{\spaceskip=\fontdimen2\font plus
\BIBentryALTinterwordstretchfactor\fontdimen3\font minus
  \fontdimen4\font\relax}
\providecommand{\BIBforeignlanguage}[2]{{%
\expandafter\ifx\csname l@#1\endcsname\relax
\typeout{** WARNING: IEEEtran.bst: No hyphenation pattern has been}%
\typeout{** loaded for the language `#1'. Using the pattern for}%
\typeout{** the default language instead.}%
\else
\language=\csname l@#1\endcsname
\fi
#2}}
\providecommand{\BIBdecl}{\relax}
\BIBdecl

\bibitem{LBP_ICIP15}
Z.~{Boulkenafet}, J.~{Komulainen}, and A.~{Hadid}, ``Face anti-spoofing based
  on color texture analysis,'' in \emph{2015 IEEE International Conference on
  Image Processing (ICIP)}, Sep. 2015, pp. 2636--2640.

\bibitem{HARALICK_BTAS16}
A.~{Agarwal}, R.~{Singh}, and M.~{Vatsa}, ``Face anti-spoofing using haralick
  features,'' in \emph{2016 IEEE 8th International Conference on Biometrics
  Theory, Applications and Systems (BTAS)}, Sep. 2016, pp. 1--6.

\bibitem{IQA_TIP14}
J.~{Galbally}, S.~{Marcel}, and J.~{Fierrez}, ``Image quality assessment for
  fake biometric detection: Application to iris, fingerprint, and face
  recognition,'' \emph{IEEE Transactions on Image Processing}, vol.~23, no.~2,
  pp. 710--724, Feb 2014.

\bibitem{IDA_TIFS15}
D.~{Wen}, H.~{Han}, and A.~K. {Jain}, ``Face spoof detection with image
  distortion analysis,'' \emph{IEEE Transactions on Information Forensics and
  Security}, vol.~10, no.~4, pp. 746--761, April 2015.

\bibitem{CNN_Arxiv14}
J.~Yang, Z.~Lei, and S.~Z. Li, ``Learn convolutional neural network for face
  anti-spoofing,'' \emph{arXiv preprint arXiv:1408.5601}, 2014.

\bibitem{BA_CVPR18}
Y.~Liu, A.~Jourabloo, and X.~Liu, ``Learning deep models for face
  anti-spoofing: Binary or auxiliary supervision,'' in \emph{Proceedings of the
  IEEE Conference on Computer Vision and Pattern Recognition (CVPR)}, June
  2018.

\bibitem{Revisiting_TBIO21}
Z.~{Yu}, X.~{Li}, J.~{Shi}, Z.~{Xia}, and G.~{Zhao}, ``Revisiting pixel-wise
  supervision for face anti-spoofing,'' \emph{IEEE Transactions on Biometrics,
  Behavior, and Identity Science}, pp. 1--1, 2021.

\bibitem{RF_Depth_BM_ECCV20}
Y.~Zhang, Z.~Yin, Y.~Li, G.~Yin, J.~Yan, J.~Shao, and Z.~Liu, ``Celeba-spoof:
  Large-scale face anti-spoofing dataset with rich annotations,'' in
  \emph{Computer Vision -- ECCV 2020}, A.~Vedaldi, H.~Bischof, T.~Brox, and
  J.-M. Frahm, Eds.\hskip 1em plus 0.5em minus 0.4em\relax Cham: Springer
  International Publishing, 2020, pp. 70--85.

\bibitem{CDCN_CVPR20}
Z.~Yu, C.~Zhao, Z.~Wang, Y.~Qin, Z.~Su, X.~Li, F.~Zhou, and G.~Zhao,
  ``Searching central difference convolutional networks for face
  anti-spoofing,'' in \emph{Proceedings of the IEEE/CVF Conference on Computer
  Vision and Pattern Recognition (CVPR)}, June 2020.

\bibitem{LBP_ECCV20}
K.-Y. Zhang, T.~Yao, J.~Zhang, Y.~Tai, S.~Ding, J.~Li, F.~Huang, H.~Song, and
  L.~Ma, ``Face anti-spoofing via disentangled representation learning,'' in
  \emph{Computer Vision -- ECCV 2020}, A.~Vedaldi, H.~Bischof, T.~Brox, and
  J.-M. Frahm, Eds.\hskip 1em plus 0.5em minus 0.4em\relax Cham: Springer
  International Publishing, 2020, pp. 641--657.

\bibitem{IDR_TIST20}
X.~Tu, Z.~Ma, J.~Zhao, G.~Du, M.~Xie, and J.~Feng, ``Learning generalizable and
  identity-discriminative representations for face anti-spoofing,'' \emph{ACM
  Transactions on Intelligent Systems and Technology (TIST)}, vol.~11, no.~5,
  pp. 1--19, 2020.

\bibitem{BM_TIFS20}
W.~{Sun}, Y.~{Song}, C.~{Chen}, J.~{Huang}, and A.~C. {Kot}, ``Face spoofing
  detection based on local ternary label supervision in fully convolutional
  networks,'' \emph{IEEE Transactions on Information Forensics and Security},
  vol.~15, pp. 3181--3196, 2020.

\bibitem{Denoising_ECCV18}
A.~Jourabloo, Y.~Liu, and X.~Liu, ``Face de-spoofing: Anti-spoofing via noise
  modeling,'' in \emph{Proceedings of the European Conference on Computer
  Vision (ECCV)}, September 2018.

\bibitem{ODPT_ECCV20}
Y.~Liu, J.~Stehouwer, and X.~Liu, ``On disentangling spoof trace for generic
  face anti-spoofing,'' in \emph{Computer Vision -- ECCV 2020}, A.~Vedaldi,
  H.~Bischof, T.~Brox, and J.-M. Frahm, Eds.\hskip 1em plus 0.5em minus
  0.4em\relax Cham: Springer International Publishing, 2020, pp. 406--422.

\bibitem{SRA_TIFS20}
A.~{Pinto}, S.~{Goldenstein}, A.~{Ferreira}, T.~{Carvalho}, H.~{Pedrini}, and
  A.~{Rocha}, ``Leveraging shape, reflectance and albedo from shading for face
  presentation attack detection,'' \emph{IEEE Transactions on Information
  Forensics and Security}, vol.~15, pp. 3347--3358, 2020.

\bibitem{SFS_TIP20}
J.~M. {Di Martino}, Q.~{Qiu}, and G.~{Sapiro}, ``Rethinking shape from shading
  for spoofing detection,'' \emph{IEEE Transactions on Image Processing},
  vol.~30, pp. 1086--1099, 2021.

\bibitem{ACMT_TIFS21}
A.~{Liu}, Z.~{Tan}, J.~{Wan}, Y.~{Liang}, Z.~{Lei}, G.~{Guo}, and S.~Z. {Li},
  ``Face anti-spoofing via adversarial cross-modality translation,'' \emph{IEEE
  Transactions on Information Forensics and Security}, pp. 1--1, 2021.

\bibitem{AMT_TMM21}
Z.~Li, H.~Li, X.~Luo, Y.~Hu, K.-Y. Lam, and A.~C. Kot, ``Asymmetric modality
  translation for face presentation attack detection,'' \emph{IEEE Transactions
  on Multimedia}, 2021.

\bibitem{DRL_TIFS20}
R.~Cai, H.~Li, S.~Wang, C.~Chen, and A.~C. Kot, ``Drl-fas: a novel framework
  based on deep reinforcement learning for face anti-spoofing,'' \emph{IEEE
  Transactions on Information Forensics and Security}, vol.~16, pp. 937--951,
  2020.

\bibitem{NASFAS_TPAMI21}
Z.~Yu, J.~Wan, Y.~Qin, X.~Li, S.~Z. Li, and G.~Zhao, ``Nas-fas: Static-dynamic
  central difference network search for face anti-spoofing,'' \emph{IEEE
  Transactions on Pattern Analysis and Machine Intelligence}, p. 1–1, 2020.

\bibitem{LWN_ICASSP20}
Z.~Yu, Y.~Qin, X.~Xu, C.~Zhao, Z.~Wang, Z.~Lei, and G.~Zhao, ``Auto-fas:
  Searching lightweight networks for face anti-spoofing,'' in \emph{ICASSP
  2020-2020 IEEE International Conference on Acoustics, Speech and Signal
  Processing (ICASSP)}.\hskip 1em plus 0.5em minus 0.4em\relax IEEE, 2020, pp.
  996--1000.

\bibitem{ANOMALY_ACCESS17}
S.~R. {Arashloo}, J.~{Kittler}, and W.~{Christmas}, ``An anomaly detection
  approach to face spoofing detection: A new formulation and evaluation
  protocol,'' \emph{IEEE Access}, vol.~5, pp. 13\,868--13\,882, 2017.

\bibitem{HYPER_ICASSP20}
Z.~{Li}, H.~{Li}, K.~Y. {Lam}, and A.~C. {Kot}, ``Unseen face presentation
  attack detection with hypersphere loss,'' in \emph{ICASSP 2020 - 2020 IEEE
  International Conference on Acoustics, Speech and Signal Processing
  (ICASSP)}, 2020, pp. 2852--2856.

\bibitem{LLIG_TIFS20}
D.~{Deb} and A.~K. {Jain}, ``Look locally infer globally: A generalizable face
  anti-spoofing approach,'' \emph{IEEE Transactions on Information Forensics
  and Security}, vol.~16, pp. 1143--1157, 2021.

\bibitem{CSAD_PR21}
S.~Fatemifar, S.~R. Arashloo, M.~Awais, and J.~Kittler, ``Client-specific
  anomaly detection for face presentation attack detection,'' \emph{Pattern
  Recognition}, vol. 112, p. 107696, 2021.

\bibitem{MK_TIFS21}
S.~R. Arashloo, ``Matrix-regularized one-class multiple kernel learning for
  unseen face presentation attack detection,'' \emph{IEEE Transactions on
  Information Forensics and Security}, vol.~16, pp. 4635--4647, 2021.

\bibitem{LDGDFR_TIFS18}
H.~Li, P.~He, S.~Wang, A.~Rocha, X.~Jiang, and A.~C. Kot, ``Learning
  generalized deep feature representation for face anti-spoofing,'' \emph{IEEE
  Transactions on Information Forensics and Security}, vol.~13, no.~10, pp.
  2639--2652, 2018.

\bibitem{MDDRL_CVPR20}
G.~Wang, H.~Han, S.~Shan, and X.~Chen, ``Cross-domain face presentation attack
  detection via multi-domain disentangled representation learning,'' in
  \emph{IEEE/CVF Conference on Computer Vision and Pattern Recognition (CVPR)},
  June 2020.

\bibitem{VSA_MM21}
J.~Wang, Z.~Zhao, W.~Jin, X.~Duan, Z.~Lei, B.~Huai, Y.~Wu, and X.~He,
  ``Vlad-vsa: Cross-domain face presentation attack detection with vocabulary
  separation and adaptation,'' in \emph{Proceedings of the 29th ACM
  International Conference on Multimedia}, 2021, pp. 1497--1506.

\bibitem{DRDG_IJCAI21}
S.~Liu, K.-Y. Zhang, T.~Yao, K.~Sheng, S.~Ding, Y.~Tai, J.~Li, Y.~Xie, and
  L.~Ma, ``Dual reweighting domain generalization for face presentation attack
  detection,'' \emph{arXiv preprint arXiv:2106.16128}, 2021.

\bibitem{LMM_AAAI20}
Y.~Qin, C.~Zhao, X.~Zhu, Z.~Wang, Z.~Yu, T.~Fu, F.~Zhou, J.~Shi, and Z.~Lei,
  ``Learning meta model for zero-and few-shot face anti-spoofing,'' in
  \emph{Proceedings of the AAAI Conference on Artificial Intelligence},
  vol.~34, no.~07, 2020, pp. 11\,916--11\,923.

\bibitem{RFM_AAAI20}
R.~Shao, X.~Lan, and P.~C. Yuen, ``Regularized fine-grained meta face
  anti-spoofing,'' in \emph{Proceedings of the AAAI Conference on Artificial
  Intelligence}, vol.~34, no.~07, 2020, pp. 11\,974--11\,981.

\bibitem{ANRL_MM21}
S.~Liu, K.-Y. Zhang, T.~Yao, M.~Bi, S.~Ding, J.~Li, F.~Huang, and L.~Ma,
  ``Adaptive normalized representation learning for generalizable face
  anti-spoofing,'' in \emph{Proceedings of the 29th ACM International
  Conference on Multimedia}, 2021, pp. 1469--1477.

\bibitem{MT_TPAMI21}
Y.~Qin, Z.~Yu, L.~Yan, Z.~Wang, C.~Zhao, and Z.~Lei, ``Meta-teacher for face
  anti-spoofing,'' \emph{IEEE Transactions on Pattern Analysis and Machine
  Intelligence}, 2021.

\bibitem{GRLMD_AAAI21}
Z.~Chen, T.~Yao, K.~Sheng, S.~Ding, Y.~Tai, J.~Li, F.~Huang, and X.~Jin,
  ``Generalizable representation learning for mixture domain face
  anti-spoofing,'' \emph{arXiv preprint arXiv:2105.02453}, 2021.

\bibitem{KSA_TIFS18}
H.~Li, W.~Li, H.~Cao, S.~Wang, F.~Huang, and A.~C. Kot, ``Unsupervised domain
  adaptation for face anti-spoofing,'' \emph{IEEE Transactions on Information
  Forensics and Security}, vol.~13, no.~7, pp. 1794--1809, 2018.

\bibitem{ADA_ICB19}
G.~Wang, H.~Han, S.~Shan, and X.~Chen, ``Improving cross-database face
  presentation attack detection via adversarial domain adaptation,'' in
  \emph{2019 International Conference on Biometrics (ICB)}.\hskip 1em plus
  0.5em minus 0.4em\relax IEEE, 2019, pp. 1--8.

\bibitem{FASDNND_SP20}
H.~Li, S.~Wang, P.~He, and A.~Rocha, ``Face anti-spoofing with deep neural
  network distillation,'' \emph{IEEE Journal of Selected Topics in Signal
  Processing}, vol.~14, no.~5, pp. 933--946, 2020.

\bibitem{USDAN-Un_PR21}
Y.~Jia, J.~Zhang, S.~Shan, and X.~Chen, ``Unified unsupervised and
  semi-supervised domain adaptation network for cross-scenario face
  anti-spoofing,'' \emph{Pattern Recognition}, vol. 115, p. 107888, 2021.

\bibitem{UDA_TIFS20}
G.~Wang, H.~Han, S.~Shan, and X.~Chen, ``Unsupervised adversarial domain
  adaptation for cross-domain face presentation attack detection,'' \emph{IEEE
  Transactions on Information Forensics and Security}, vol.~16, pp. 56--69,
  2020.

\bibitem{SDA_TIFS15}
J.~Yang, Z.~Lei, D.~Yi, and S.~Z. Li, ``Person-specific face antispoofing with
  subject domain adaptation,'' \emph{IEEE Transactions on Information Forensics
  and Security}, vol.~10, no.~4, pp. 797--809, 2015.

\bibitem{DGP_ICASSP20}
A.~Mohammadi, S.~Bhattacharjee, and S.~Marcel, ``Domain adaptation for
  generalization of face presentation attack detection in mobile settengs with
  minimal information,'' in \emph{ICASSP 2020-2020 IEEE International
  Conference on Acoustics, Speech and Signal Processing (ICASSP)}.\hskip 1em
  plus 0.5em minus 0.4em\relax IEEE, 2020, pp. 1001--1005.

\bibitem{OCA-FAS_NC20}
Y.~Qin, W.~Zhang, J.~Shi, Z.~Wang, and L.~Yan, ``One-class adaptation face
  anti-spoofing with loss function search,'' \emph{Neurocomputing}, vol. 417,
  pp. 384--395, 2020.

\bibitem{MP_TIFS22}
R.~Cai, Z.~Li, R.~Wan, H.~Li, Y.~Hu, and A.~C. Kot, ``Learning meta pattern for
  face anti-spoofing,'' \emph{IEEE Transactions on Information Forensics and
  Security}, vol.~17, pp. 1201--1213, 2022.

\bibitem{OULU}
Z.~Boulkenafet, J.~Komulainen, L.~Li, X.~Feng, and A.~Hadid, ``Oulu-npu: A
  mobile face presentation attack database with real-world variations,'' in
  \emph{2017 12th IEEE International Conference on Automatic Face \& Gesture
  Recognition (FG 2017)}.\hskip 1em plus 0.5em minus 0.4em\relax IEEE, 2017,
  pp. 612--618.

\bibitem{competition}
Z.~Boulkenafet, J.~Komulainen, Z.~Akhtar, A.~Benlamoudi, D.~Samai, S.~E.
  Bekhouche, A.~Ouafi, F.~Dornaika, A.~Taleb-Ahmed, L.~Qin \emph{et~al.}, ``A
  competition on generalized software-based face presentation attack detection
  in mobile scenarios,'' in \emph{2017 IEEE International Joint Conference on
  Biometrics (IJCB)}.\hskip 1em plus 0.5em minus 0.4em\relax IEEE, 2017, pp.
  688--696.

\bibitem{US_CVPR20}
P.~Bergmann, M.~Fauser, D.~Sattlegger, and C.~Steger, ``Uninformed students:
  Student-teacher anomaly detection with discriminative latent embeddings,'' in
  \emph{Proceedings of the IEEE/CVF Conference on Computer Vision and Pattern
  Recognition}, 2020, pp. 4183--4192.

\bibitem{VGG_Arxiv14}
K.~Simonyan and A.~Zisserman, ``Very deep convolutional networks for
  large-scale image recognition,'' \emph{arXiv preprint arXiv:1409.1556}, 2014.

\bibitem{DeepPixBis_ICB19}
A.~George and S.~Marcel, ``Deep pixel-wise binary supervision for face
  presentation attack detection,'' in \emph{2019 International Conference on
  Biometrics (ICB)}.\hskip 1em plus 0.5em minus 0.4em\relax IEEE, 2019, pp.
  1--8.

\bibitem{SNFS_Axiv19}
T.~Dettmers and L.~Zettlemoyer, ``Sparse networks from scratch: Faster training
  without losing performance,'' \emph{arXiv preprint arXiv:1907.04840}, 2019.

\bibitem{CASIA}
Z.~Zhang, J.~Yan, S.~Liu, Z.~Lei, D.~Yi, and S.~Z. Li, ``A face antispoofing
  database with diverse attacks,'' in \emph{2012 5th IAPR International
  Conference on Biometrics (ICB)}.\hskip 1em plus 0.5em minus 0.4em\relax IEEE,
  2012, pp. 26--31.

\bibitem{MSU}
D.~Wen, H.~Han, and A.~K. Jain, ``Face spoof detection with image distortion
  analysis,'' \emph{IEEE Transactions on Information Forensics and Security},
  vol.~10, no.~4, pp. 746--761, 2015.

\bibitem{IDIAP}
I.~Chingovska, A.~Anjos, and S.~Marcel, ``On the effectiveness of local binary
  patterns in face anti-spoofing,'' in \emph{2012 BIOSIG-Proceedings of the
  International Conference of Biometrics Special Interest Group
  (BIOSIG)}.\hskip 1em plus 0.5em minus 0.4em\relax IEEE, 2012, pp. 1--7.

\bibitem{SVDD}
D.~M. Tax and R.~P. Duin, ``Support vector data description,'' \emph{Machine
  learning}, vol.~54, no.~1, pp. 45--66, 2004.

\bibitem{GMM}
A.~P. Dempster, N.~M. Laird, and D.~B. Rubin, ``Maximum likelihood from
  incomplete data via the em algorithm,'' \emph{Journal of the Royal
  Statistical Society: Series B (Methodological)}, vol.~39, no.~1, pp. 1--22,
  1977.

\bibitem{ADAM_ICLR15}
D.~P. Kingma and J.~Ba, ``Adam: A method for stochastic optimization,''
  \emph{arXiv preprint arXiv:1412.6980}, 2014.

\bibitem{ADDA_CVPR17}
E.~Tzeng, J.~Hoffman, K.~Saenko, and T.~Darrell, ``Adversarial discriminative
  domain adaptation,'' in \emph{Proceedings of the IEEE Conference on Computer
  Vision and Pattern Recognition}, 2017, pp. 7167--7176.

\bibitem{DRCN_ECCV16}
M.~Ghifary, W.~B. Kleijn, M.~Zhang, D.~Balduzzi, and W.~Li, ``Deep
  reconstruction-classification networks for unsupervised domain adaptation,''
  in \emph{European Conference on Computer Vision}.\hskip 1em plus 0.5em minus
  0.4em\relax Springer, 2016, pp. 597--613.

\bibitem{DUPGAN_CVPR18}
L.~Hu, M.~Kan, S.~Shan, and X.~Chen, ``Duplex generative adversarial network
  for unsupervised domain adaptation,'' in \emph{Proceedings of the IEEE
  Conference on Computer Vision and Pattern Recognition}, 2018, pp. 1498--1507.

\bibitem{DTN_CVPR19}
Y.~Liu, J.~Stehouwer, A.~Jourabloo, and X.~Liu, ``Deep tree learning for
  zero-shot face anti-spoofing,'' in \emph{Proceedings of the IEEE/CVF
  Conference on Computer Vision and Pattern Recognition}, 2019, pp. 4680--4689.

\end{thebibliography}

\end{document}